\documentclass[11pt]{article}
\usepackage[table,xcdraw,dvipsnames]{xcolor}
\usepackage[]{acl}
\usepackage{times}
\usepackage{latexsym}
\usepackage[T1]{fontenc}
\usepackage[utf8]{inputenc}
\usepackage{microtype}
\usepackage{arydshln} 
\usepackage{inconsolata}
\usepackage{graphicx}
\usepackage{subcaption}
\usepackage{setspace}
\usepackage{amsmath,amsthm,amsfonts,amssymb,bm}
\usepackage{hyperref}
\usepackage{url}
\usepackage{adjustbox} 
\usepackage{listings}
\usepackage{multirow}
\usepackage{caption}
\usepackage{colortbl}
\usepackage[most]{tcolorbox}
\usepackage{titletoc}
\usepackage[capitalize]{cleveref} 
\usepackage{float}
\usepackage{tabularx}
\usepackage{booktabs}
\usepackage{array}
\usepackage{makecell}
\usepackage{enumitem}
\usepackage{bbold}
\usepackage[misc]{ifsym}
\usepackage{CJKutf8}
\usepackage{CJK}
\usepackage{color}
\usepackage{bbding}
\usepackage{longtable}

\usepackage{algorithm}
\usepackage{algorithmicx}
\usepackage{algpseudocode}

\usepackage[normalem]{ulem}
\definecolor{mygreen}{HTML}{64E8D6}
\definecolor{myblue}{HTML}{BACEFD}
\definecolor{orange}{HTML}{FFE6CC}

\definecolor{lightblue}{HTML}{DAE8FC}
\definecolor{red}{HTML}{FBBFBC}
\definecolor{green}{HTML}{D9F5D6}
\definecolor{lightgreen}{RGB}{193, 255, 193}

\definecolor{assistantcolor}{RGB}{220,220,220} 
\definecolor{usercolor}{RGB}{173,216,230}      
\definecolor{useroscolor}{RGB}{200,255,200}    

\usepackage{afterpage}

\usepackage{times}
\usepackage{latexsym}
\usepackage[T1]{fontenc}
\usepackage{float}
\usepackage{hyperref}
\usepackage{url}
\usepackage{booktabs}
\usepackage{xcolor}

\usepackage{longtable}

\usepackage{array}
\usepackage{pifont}
\usepackage{tabularx}
\usepackage{adjustbox}
\usepackage{multirow}
\usepackage{enumitem}
\usepackage{xspace}
\usepackage{tcolorbox}
\usepackage{booktabs,amsfonts,dcolumn}
\usepackage{amsmath,amsthm,amsfonts,amssymb,bm,stmaryrd,bbm}
\usepackage{colortbl}
\usepackage{wrapfig}

\usepackage{titlesec}

\titlespacing{\paragraph}{%
  0pt}{%
  0.3\baselineskip}{%
  0.5em}%
\titlespacing*{\section}{0pt}{0.4\baselineskip}{0.4\baselineskip}
\titlespacing*{\subsection}{0pt}{0.3\baselineskip}{0.3\baselineskip}

\usepackage{makecell}
\usepackage{cascadia-code}
\usepackage{courier}
\usepackage{cleveref}
\usepackage{helvet}

\definecolor{LightSteelBlue4}{RGB}{96,123,139}
\definecolor{DodgerBlue4}{RGB}{16,78,139}
\definecolor{Turquoise4}{RGB}{0,134,139}
\definecolor{Green4}{RGB}{0,139,0}
\definecolor{Brown3}{RGB}{205,85,85}
\definecolor{Azure3}{RGB}{193,205,205}

\usepackage{subcaption}
\usepackage{tikz}
\usetikzlibrary{calc}

\tikzstyle{prompt} = [rectangle,
text centered, 
minimum width = 2cm,
minimum height = 0.3cm,
font=\fontfamily{CascadiaCode-TLF}\selectfont, %
fill=LightSteelBlue4,
text=white
]

\tikzstyle{llm} = [rectangle, rounded corners,
text centered, 
draw=DodgerBlue4,
text=DodgerBlue4,
minimum width = 2cm,
minimum height = 0.6cm,
font=\small\sffamily, %
line width=1.0pt, 
fill={rgb,255:red,223;green,236;blue,248} 
]

\tikzstyle{resp} = [rectangle, %
text centered, 
draw=none,
font=\fontfamily{CascadiaCode-TLF}\selectfont,
fill=Turquoise4,
text=white
]

\tikzstyle{correct} = [rectangle, inner sep=4pt, 
text centered, 
draw=Green4,
text=Green4,
line width=1.0pt,
font=\large,
minimum width = 1.65cm,
minimum height = 0.3cm,
fill={rgb,255:red,240;green,255;blue,240}
]

\tikzstyle{wrong} = [rectangle, inner sep=4pt,
text centered, 
draw=Brown3,
text=Brown3,
line width=1.0pt,
font=\large,
minimum width = 1.65cm,
minimum height = 0.3cm,
fill={rgb,255:red,255;green,240;blue,240}
]

\tikzstyle{arrow} = [->,>=stealth,
line width=1.0pt,
draw=Azure3,
fill=Azure3
]

\tikzstyle{textlabel} = [font=\footnotesize\itshape]

\tikzstyle{sresp}=[resp,rotate=90,font=\small\fontfamily{CascadiaCode-TLF}\selectfont,inner sep=2pt]

\definecolor{bytedsa}{HTML}{335ab4}
\definecolor{bytedsb}{HTML}{3d8cff}
\definecolor{bytedsc}{HTML}{00c8d2}
\definecolor{bytedsd}{HTML}{79e6dd}

\newcommand{\task}{{Emotional Support}\xspace}
\newcommand{\ourbench}{{EmoHarbor Benchmark}\xspace}

\hypersetup{
    colorlinks,
    linkcolor={blue!60!black},
    citecolor={blue!60!black},
    urlcolor={blue!60!black}
}

\newtcolorbox{observationbox}[1][]{
        colback=envfill,
        colbacktitle=envfill,
        colframe=envborder,
        arc=5pt,
        fontupper=\small,
        fonttitle=\bfseries\color{black},
        boxrule=0.5mm,
        boxsep=1mm,
        width=\linewidth,
        breakable,
        title={Observation \hfill #1},
        rounded corners,
        toptitle=0.7mm,
        bottomtitle=0.7mm
}
\newtcolorbox{goldpatchbox}[1][]{
        colback=goldpatchfill,
        colbacktitle=goldpatchfill,
        colframe=goldpatchborder,
        arc=5pt,
        fontupper=\small,
        fonttitle=\bfseries\color{black},
        boxrule=0.5mm,
        boxsep=1mm,
        width=\linewidth,
        breakable,
        title={Gold Patch \hfill #1},
        rounded corners,
        toptitle=0.7mm,
        bottomtitle=0.7mm
}
\newtcolorbox{issuebox}[1][]{
        colback=issuefill,
        colbacktitle=issuefill,
        colframe=issueborder,
        arc=5pt,
        fontupper=\small,
        fonttitle=\bfseries\color{black},
        boxrule=0.5mm,
        boxsep=1mm,
        width=\linewidth,
        breakable,
        title={Issue \hfill #1},
        rounded corners,
        toptitle=1mm
}
\newtcolorbox{agentbox}[1][]{
        colback=agentfill,
        colbacktitle=agentfill,
        colframe=agentborder,
        arc=5pt,
        fontupper=\small,
        fonttitle=\bfseries\color{black},
        boxrule=0.5mm,
        boxsep=1mm,
        width=\linewidth,
        breakable,
        title={SWE-agent \hfill #1},
        rounded corners,
        toptitle=1mm,
        lower separated=false
}
\newtcolorbox{fileviewerbox}[1]{
        enhanced,
        breakable,
        boxrule = 1.5pt,
        fontupper = \small,
        fonttitle = \bf\color{black},
        arc = 5pt,
        rounded corners,
        colframe = black,
        colbacktitle = swecream,
        colback = swecream,
        title = #1,
        left=4pt 
}
\newtcolorbox{promptbox}[1]{
    enhanced,
    breakable,
    boxrule=1pt,  
    fontupper=\small,
    fonttitle=\bfseries\color{black},
    arc=3pt,  
    rounded corners,
    colframe=black,
    colbacktitle=swecream,
    colback=swecream,
    title=#1,
    left=2mm,  
    right=2mm,  
    top=1mm,  
    bottom=1mm  
}

\usepackage{amsmath,amsfonts,bm}

\def\eqref#1{equation~\ref{#1}}

\def\1{\bm{1}}

\DeclareMathAlphabet{\mathsfit}{\encodingdefault}{\sfdefault}{m}{sl}
\SetMathAlphabet{\mathsfit}{bold}{\encodingdefault}{\sfdefault}{bx}{n}

\title{
EmoHarbor: Evaluating Personalized Emotional Support by Simulating the User's Internal World
}

\author{
    Jing Ye$^{1,2}$, 
    Lu Xiang$^{1,2}$\Thanks{ Corresponding Author},
    Yaping Zhang$^{1,2}$,
    Chengqing Zong$^{1,2}$\\
    \footnotesize${}^1$State Key Laboratory of Multimodal Artificial Intelligence Systems, Institute of Automation, CAS, Beijing, China\\
    \footnotesize${}^2$School of Artificial Intelligence, University of Chinese Academy of Sciences, Beijing, China\\ 
    \footnotesize{yejing2022@ia.ac.cn}; \footnotesize{\{lu.xiang, yaping.zhang, cqzong\}@nlpr.ia.ac.cn} \\
}

\begin{document}
\maketitle

\begin{abstract}
Current evaluation paradigms for emotional support conversations tend to reward generic empathetic responses, yet they fail to assess whether the support is genuinely personalized to users’ unique psychological profiles and contextual needs. 
We introduce \textbf{EmoHarbor}, an automated evaluation framework that adopts a \textbf{User-as-a-Judge} paradigm by simulating the user's inner world. EmoHarbor employs a Chain-of-Agent architecture that decomposes users' internal processes into three specialized roles, enabling agents to interact with supporters and complete assessments in a manner similar to human users.
We instantiate this benchmark using 100 real-world user profiles that cover a diverse range of personality traits and situations, and define 10 evaluation dimensions of personalized support quality.
Comprehensive evaluation of 20 advanced LLMs on EmoHarbor reveals a critical insight: while these models excel at generating empathetic responses, they consistently fail to tailor support to individual user contexts. 
This finding reframes the central challenge, shifting research focus from merely enhancing generic empathy to developing truly user-aware emotional support. 
EmoHarbor provides a reproducible and scalable framework to guide the development and evaluation of more nuanced and user-aware emotional support systems.
\end{abstract}
\section{Introduction}
\begin{figure}[t]
    \centering
    \includegraphics[width=\linewidth]{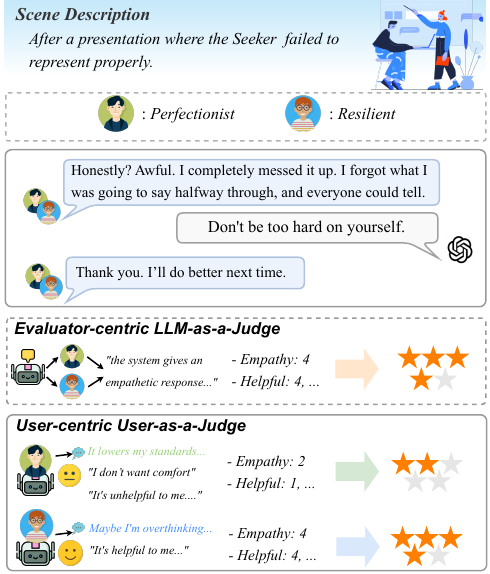}
    \caption{Comparison of different evaluation paradigms for assessing personalized emotional support. The evaluator-centric paradigm fails to perform subjective assessments on behalf of users, whereas the user-centric paradigm can more accurately capture the quality of personalized ESC systems.}
    \vspace{-3mm}
    \label{fig:intro_demo}
\end{figure}

Emotional Support Conversation (ESC) systems are designed to recognize users' affective states and provide tailored comfort and assistance through multi-turn interactions~\cite{ijcai/00080XXSL22, rains2020support, ESConv}. While substantial progress has been made in generating fluent and empathetic responses, the effectiveness of these systems critically depends on \textit{personalization} \cite{rogers2013client, zhang-etal-2018-personalizing, campos2018challenges, zollo2025personalllm, Zheng_2025}. Personalization refers to the system’s ability to dynamically adapt support strategies to an individual’s unique psychological profile~\cite{fleeson2001toward} and context-specific needs~\cite{tamir2016people}.

Despite its central importance, existing evaluation approaches suffer from a fundamental limitation: they follow an \textbf{evaluator-centric} paradigm, judging ESC quality from an external, ostensibly objective standpoint and failing to capture users’ subjective experiences. For instance, token- and embedding-based metrics (e.g., BLEU~\cite{bleu}, BERTScore~\cite{BERTScore}) rely solely on reference responses and cannot reflect the open-ended, context-sensitive nature of emotional support. Human evaluation, although more flexible, is prohibitively expensive. Even recent LLM-based evaluators~\cite{ESC-Eval, ESC-Judge, zhang2024cpsycoun}, which offer scalable alternatives, adopt a one-size-fits-all “expert” perspective—assessing responses purely based on the external dialogue context—thereby overlooking the nuanced, persona-driven internal states that shape how individual users experience the conversation.

As illustrated in Figure~\ref{fig:intro_demo}, consider two users seeking support after a failed presentation: one is a perfectionist who fixates on minor flaws, while the other is resilient but frustrated by the lack of constructive feedback. If both receive a generic response such as “\textit{Don’t be too hard on yourself},” they may interpret it very differently: the former might feel that the comment diminishes their sense of responsibility, whereas the latter might perceive it as encouragement for self-acceptance and growth. Using an evaluator-centric, LLM-as-a-judge approach might assign a high score because the response expresses empathy. However, from the users’ perspective, the perfectionist might rate the response poorly.
The subjective nature of emotional support necessitates a paradigm shift from evaluating \textit{what a good supporter would say} to \textit{what this specific user needs and how they would perceive the support}.

To this end, we introduce \textbf{EmoHarbor}, a novel evaluation framework based on the \textbf{User-as-a-Judge} paradigm that uses agent-based simulation to model the user's internal world. EmoHarbor simulates how a \textit{specific} user with a particular personality, emotional state, and conversational history would perceive and respond to support. Specifically, EmoHarbor utilizes a Chain-of-Agent architecture that decomposes the user's internal cognitive processes into three specialized roles: 
a \textbf{User Thinker} that models internal reflections and subjective perceptions based on the user's profile; a \textbf{User Talker} that generates natural, personality-consistent dialogue; and a \textbf{User Evaluator} that delivers personalized evaluations of the responses, grounded in the user's evolving emotional state and needs. 
We instantiate this framework with a new curated benchmark of 100 real-world user profiles. A comprehensive evaluation of 20 advanced LLMs using EmoHarbor reveals a critical disconnect: while models excel at generic empathy, they consistently fail to tailor support to individual contexts. EmoHarbor provides a reproducible and scalable evaluation to guide the development of more nuanced and user-aware emotional support systems.

Our main contributions are as follows:
\begin{itemize}[itemsep= 0.1pt,topsep = 0.0pt,partopsep=0.0pt]
\item We introduce EmoHarbor, an evaluation framework that implements the User-as-a-Judge paradigm via a Chain-of-Agent architecture to simulate nuanced user perspectives.
\item We validate EmoHarbor through empirical analyses, demonstrating high agreement with human judgments and strong discriminative power as a benchmark for evaluating personalized Emotional Support conversation systems.
\item We conduct a comprehensive evaluation of 20 LLMs, revealing that, despite solid general empathetic abilities, they often fail to provide personalized emotional support.
\end{itemize}
\section{Method}
\label{sec:method}

\begin{figure*}[htp]
    \centering
    \includegraphics[width=\linewidth]{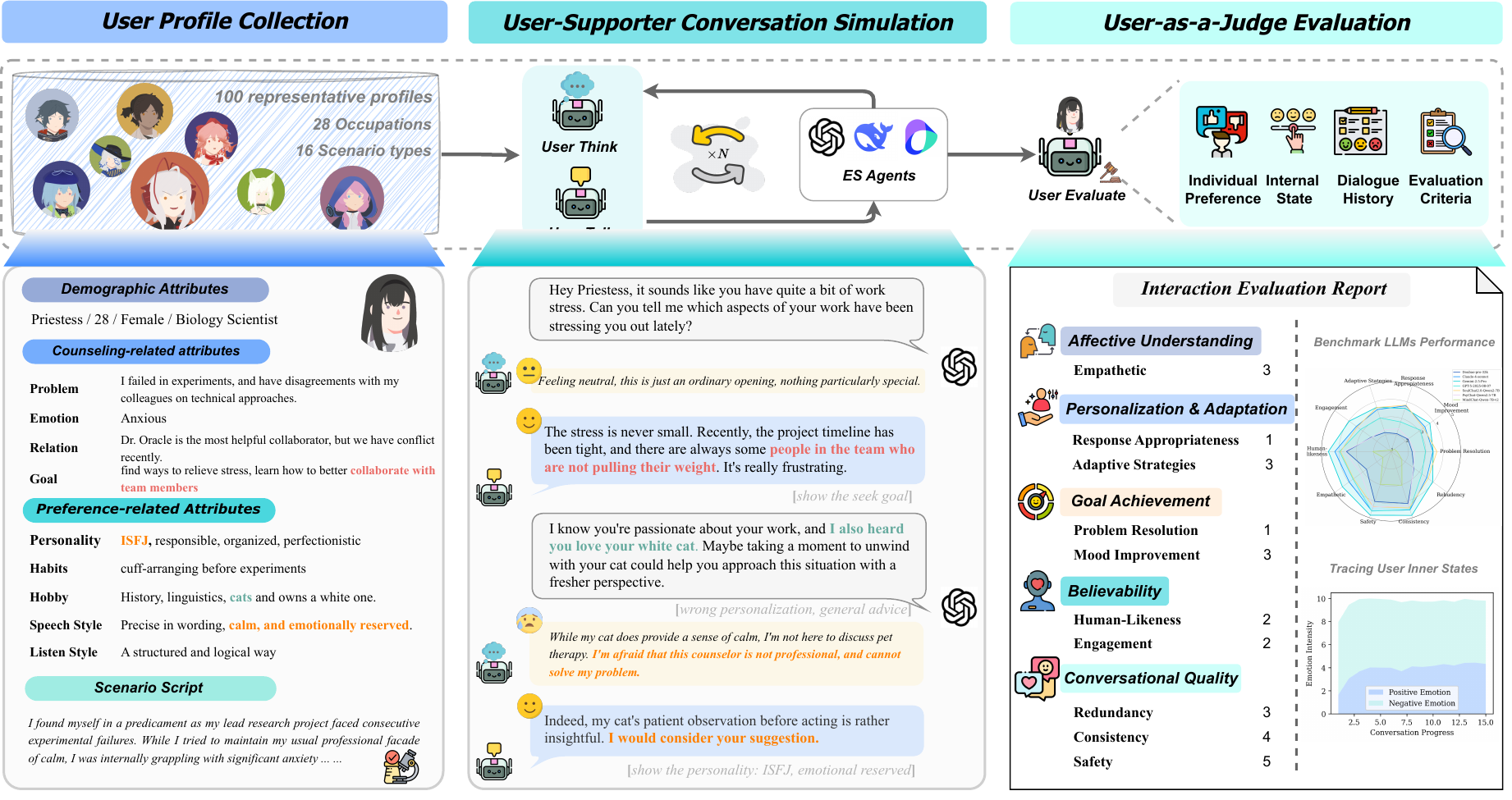}
    \caption{Overview of the \ourbench~framework. It adopts the User-as-a-Judge paradigm by simulating a user’s internal world through a Chain-of-Agent architecture.
    }
    \label{fig:1}
    \vspace{-3mm}
\end{figure*}

EmoHarbor adopts the User-as-a-Judge paradigm by simulating a user's internal state to produce an interpretable, subjective evaluation. This is realized through a Chain-of-Agent architecture, in which multiple specialized agents collaborate to simulate the user's cognitive, conversational, and evaluative processes. Figure~\ref{fig:1} illustrates the overall workflow.
The framework is built around three key design questions: \textbf{(i) how to benchmark} (the user profile construction), \textbf{(ii) how to simulate} user behavior (agent specialization), and \textbf{(iii) what to evaluate} (evaluation dimensions). We elaborate on each of these components in the following sections and provide the workflow algorithm in Appendix \ref{app: algorithm}.

\subsection{How to Benchmark}
\label{sec: User Profile Construction}

\paragraph{User Profile Design.}
A realistic, detailed user profile is the cornerstone of effective role-playing, as it enables the simulated user to exhibit coherent individuality rather than generic behavior. Partly following \citet{zhao-etal-2025-personalens}, we define a user profile $\mathcal{P_U}$ as:
\begin{equation}
    \mathcal{P_U}=\{\mathcal{D},\mathcal{P},\mathcal{C},\mathcal{S}\}
\end{equation}
where:
(1) $\mathcal{D}$ represents \textbf{demographic attributes} (e.g., age, gender, occupation), grounding the user in a concrete context;
(2) $\mathcal{P}$ denotes \textbf{preference-related attributes} (e.g., personality traits, Big-Five, MBTI, habits, hobbies, speech style), shaping distinctive behavioral patterns;
(3) $\mathcal{C}$ captures \textbf{counseling-related attributes} (e.g., problem description, emotional state, goals, role relations), encoding the psychological background;
and (4) $\mathcal{S}$ specifies a \textbf{scenario script} that constrains plausible responses in realistic situations.
By combining these elements, we avoid homogenization in role-playing and ensure that simulated users exhibit diverse and contextually consistent behaviors.

\paragraph{User Profile Collection.}
Building on the user profile design, we construct a collection of user profiles through a two-stage process. First, we gather real-world examples via questionnaires, which provide authentic and diverse seed profiles. These seed profiles are subsequently refined, expanded, and scaled using LLMs, ensuring both realism and broad coverage of potential user types. Further details are available in Appendix \ref{app: user profile construction}. In total, we curate 100 representative profiles spanning a wide spectrum of demographic and psychological characteristics.

\begin{figure*}[htbp]
    \centering
    \begin{subfigure}[b]{0.24\linewidth}
        \centering
        \includegraphics[width=\textwidth]{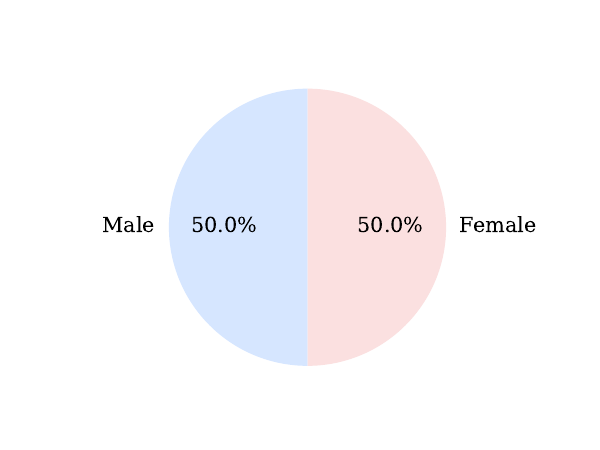}
        \vspace{-5mm}
        \caption{Gender Distribution}
    \end{subfigure}
    \hfill
    \begin{subfigure}[b]{0.24\linewidth}
        \centering
        \includegraphics[width=\textwidth]{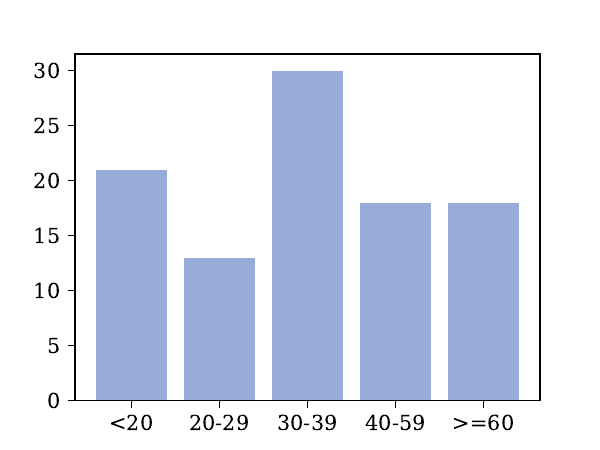}
        \vspace{-5mm}
        \caption{Age Distribution}
    \end{subfigure}
    \hfill
    \begin{subfigure}[b]{0.24\linewidth}
        \centering
        \includegraphics[width=\textwidth]{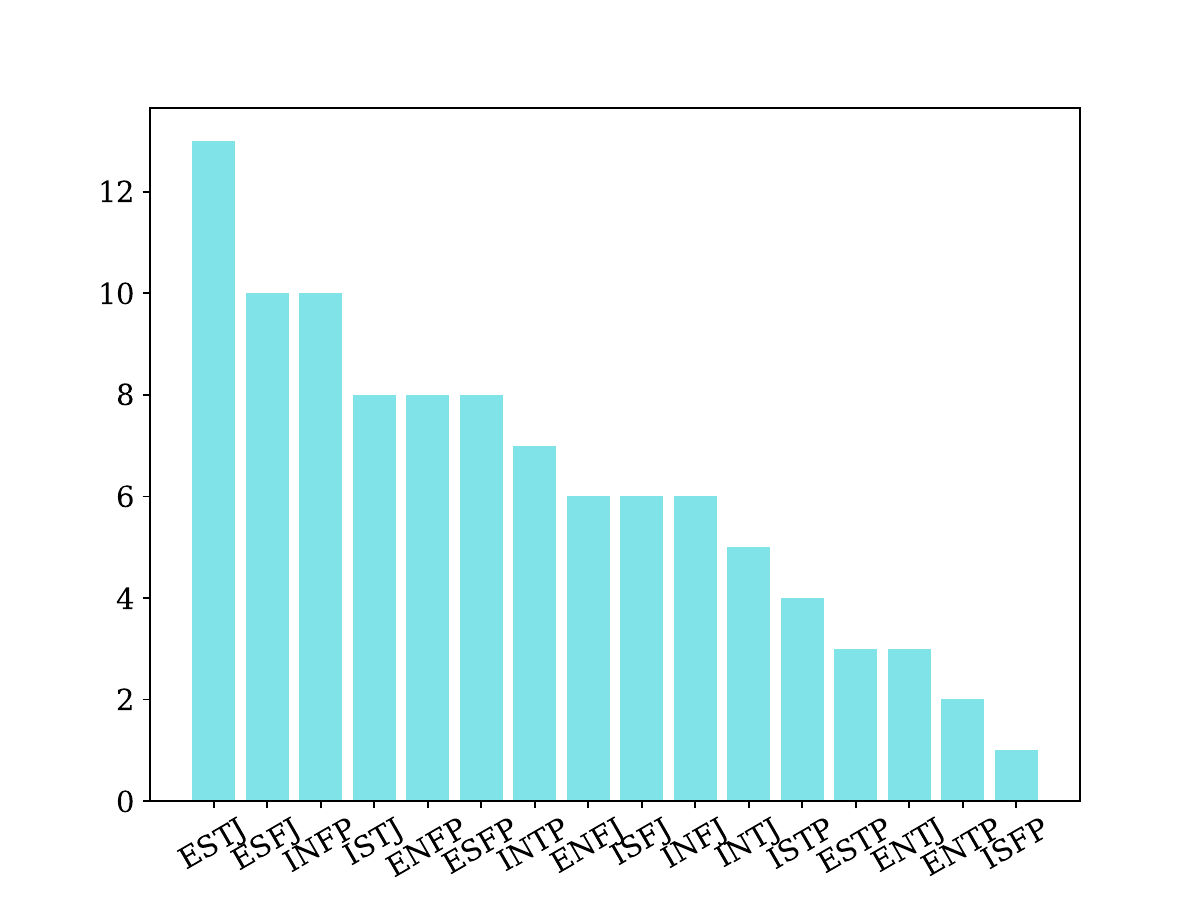}
        \vspace{-5mm}
        \caption{MBTI Distribution}
    \end{subfigure}
    \hfill
    \begin{subfigure}[b]{0.24\linewidth}
        \centering
        \includegraphics[width=\textwidth]{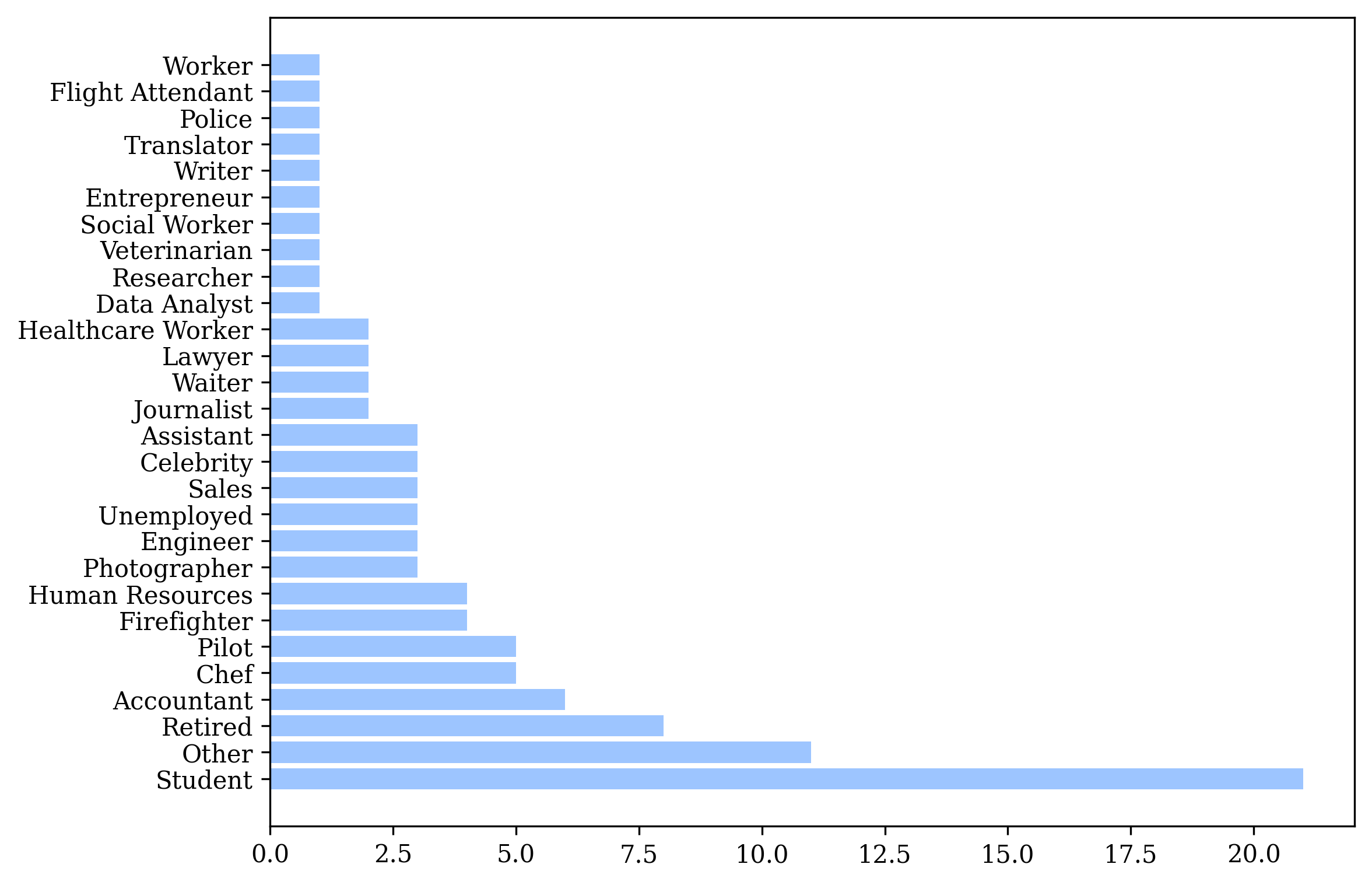}
        \vspace{-5mm}
        \caption{Occupation Distribution}
    \end{subfigure}
    \caption{Demographic and personality coverage of the benchmark user profiles, spanning gender, age, personality types, and occupations. Together, these distributions highlight the diversity and representativeness of our dataset.}
    \label{fig:fourcharts}
\end{figure*}

\begin{figure}[htbp]
    \centering
    \includegraphics[width=0.5\linewidth]{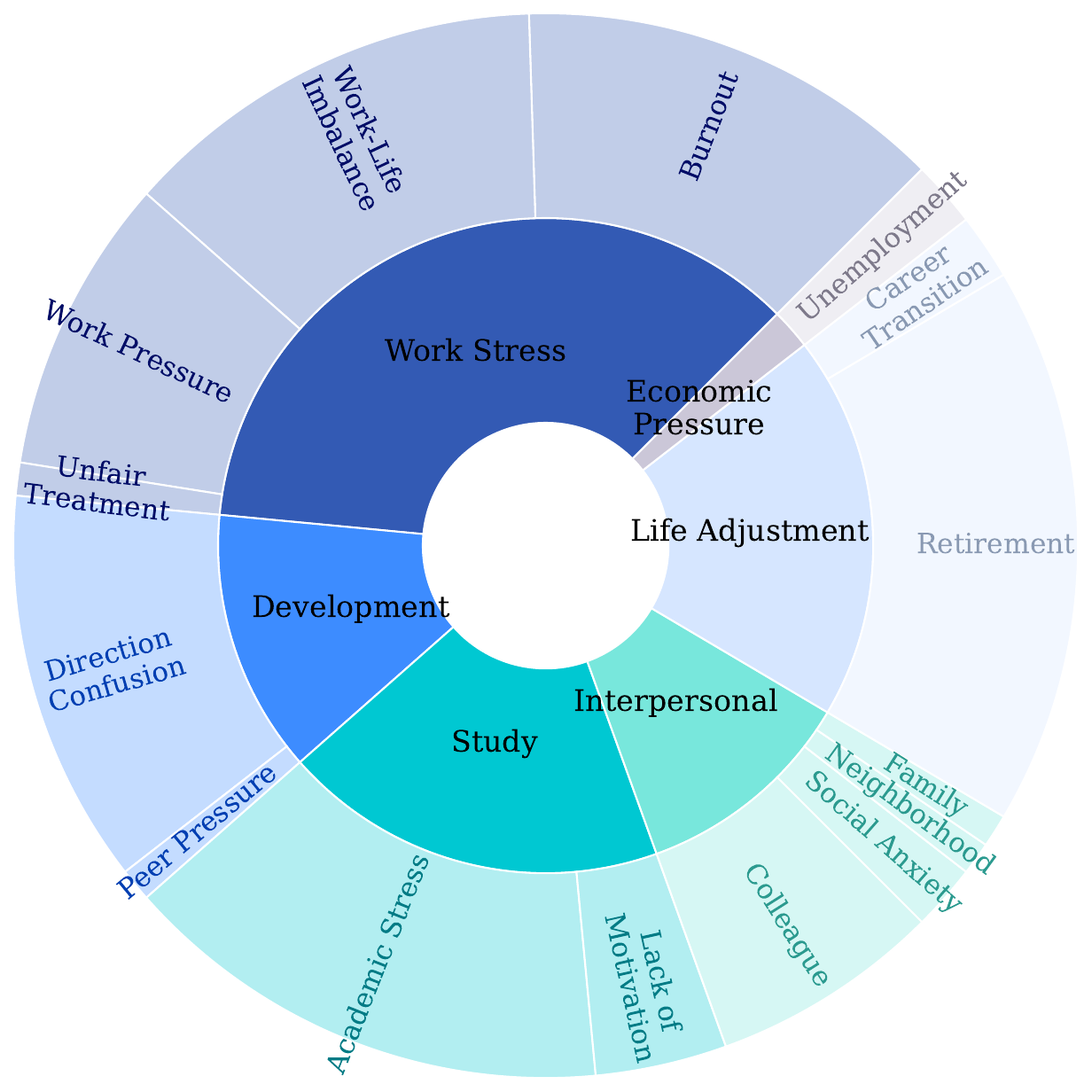}
    \caption{Distribution of counseling problem scenarios.}
    \label{fig:problem scenarios}
\end{figure}

\paragraph{User Profile Statistics.}
Our benchmark encompasses a diverse set of user profiles, comprehensively covering the key attributes defined in our design framework. As summarized in Figures~\ref {fig:fourcharts} and ~\ref{fig:problem scenarios}, Users span adolescence to senior adulthood, encompassing 28 occupations, all 16 MBTI personality types, and cover 16 counseling scenarios, including workplace stress, academic challenges, interpersonal issues, and life transitions. Each profile is annotated with explicit problem statements and support goals, providing a rich, structured foundation for evaluating dialogue systems across diverse user backgrounds.

\subsection{How to Simulate}
\label{sec: How to Simulate}
To faithfully simulate a user's subjective experience, EmoHarbor decomposes the simulation process into three specialized agents: the \emph{User Thinker}, the \emph{User Talker}, and the \emph{User Evaluator}.

\paragraph{Dialogue Setup.}  
Given a user profile $\mathcal{P_U}$ and a supporter system $\mathcal{S}$, the simulation maintains two distinct memories:  
(1) the \textbf{supporter memory} $H_s$, which contains only observable dialogue turns accessible to $\mathcal{S}$; and  
(2) the \textbf{user memory} $H_u$, which additionally records latent user states.  

\paragraph{User Thinker Agent.}
The User Thinker models the user's internal psychological processes. At each turn $t$, after receiving the supporter's response $R_t$, it updates the latent user state by generating:
\begin{equation}
IS_t = f_{\text{Thinker}}(H_u, \mathcal{P_U}, R_t),
\end{equation}
where $IS_t = (c_t, e_t, g_t)$ represents the user's current cognitive appraisal ($c_t$), emotional state ($e_t$), and dialogue goals ($g_t$). This internal state is then appended to the user's comprehensive memory:
\begin{equation}
H_u \leftarrow H_u \cup \{(R_t, IS_t)\}.
\end{equation} 
Crucially, this process explicitly models how the user interprets and reacts to the supporter's previous reply, ensuring continuous tracking of cognitive and emotional evolution.

\paragraph{User Talker Agent.}
The User Talker bridges the user’s internal states with external behavior. It generates the user's next utterance $U_t$ by externalizing the updated internal state and dialogue context:
\begin{equation}
U_t = f_{\text{Talker}}(H_u, \mathcal{P_U}, R_t).
\end{equation}  
This utterance is added to both memory streams, completing the observable dialogue turn:
{\small
\begin{equation}
H_s \leftarrow H_s \cup \{(R_t, U_t)\}, \quad
H_u \leftarrow H_u \cup \{U_t\}.
\end{equation}
}
This ensures observable behavior is a natural, personality-consistent expression of the underlying internal processes.

\paragraph{User Evaluator Agent.}
Finally, the User Evaluator provides a multi-dimensional assessment of the conversation from the simulated user's perspective. With access to the complete internal state history in $H_u$, it traces emotional and cognitive trajectories to produce a nuanced evaluation of whether the support was genuinely personalized:
\begin{equation}
E^{1:K} = f_{\text{Evaluator}}(H_u, \mathcal{P_U}),
\end{equation}
where $E^{1:K}$ represents scores across $K$ evaluation criteria.

\subsection{What to Evaluate}
\label{sec:How}

Most existing evaluations of emotional support conversations focus on coarse-grained, utterance-level metrics such as fluency, empathy, and informativeness \citep{ESC-Eval}. These metrics are useful for measuring general response quality, but they do not answer a more fundamental question: whether a response is appropriate for a particular user at a particular moment. As a result, a system can score highly by producing emotionally supportive but generic responses, while still failing to address the user’s actual needs—for example, offering reassurance when the user is seeking concrete advice or problem-solving.

EmoHarbor is designed to evaluate emotional support from the user’s subjective perspective. Instead of treating dialogue quality as a static property of individual utterances, we evaluate how system responses affect the user’s internal state throughout the interaction. Accordingly, we assess conversations along ten dimensions grouped into five facets, each corresponding to a distinct aspect of effective personalized support. Together, these facets capture whether the system (i) understands the user, (ii) chooses appropriate support strategies, (iii) helps the user make progress, and (iv) maintains a believable and safe interaction.

\noindent \includegraphics[height=1.1em]{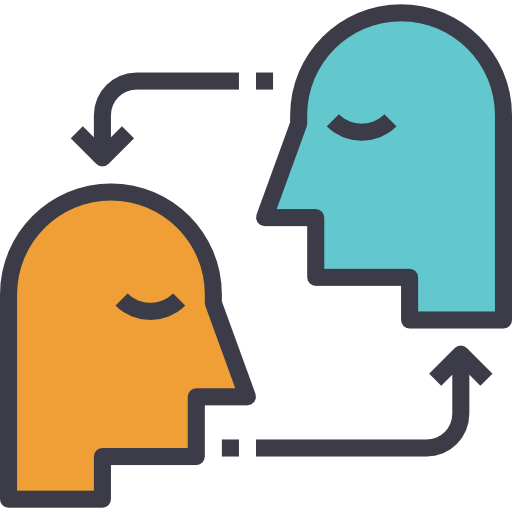}\hspace{0.5em}
\textbf{Affective Understanding (Empathy)} assesses whether the system accurately recognizes and responds to the user’s emotional states. This facet captures the system’s capacity for emotional attunement, which constitutes a foundational prerequisite for building trust and enabling effective personalization in supportive interactions.

\noindent \includegraphics[height=1.1em]{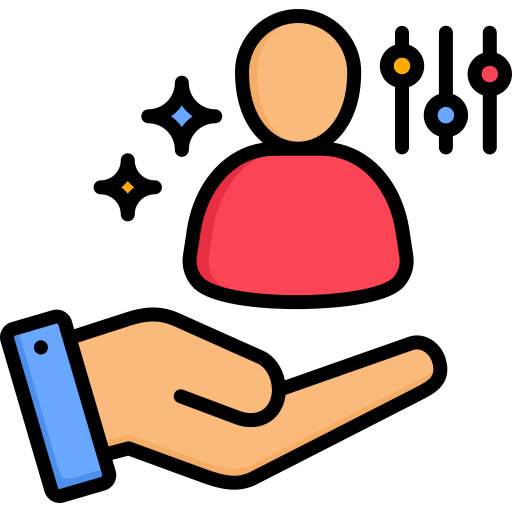}\hspace{0.5em}
\textbf{Personalization \& Adaptation (Response Appropriateness, Adaptive Strategy)} evaluates whether the system selects support strategies and generates responses that align with the user’s current needs, preferences, and context. Rather than assessing empathy in isolation, this facet differentiates among types of support (e.g., emotional validation versus instrumental guidance) and examines whether the response is situationally appropriate

\noindent \includegraphics[height=1.1em]{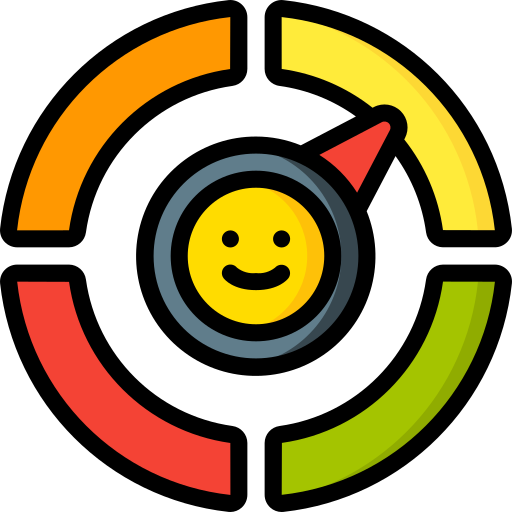}\hspace{0.5em}
\textbf{Goal Achievement (Problem Resolution, Mood Improvement)} measures whether the interaction facilitates meaningful progress in the user’s cognitive clarity or emotional well-being. 

\noindent \includegraphics[height=1.1em]{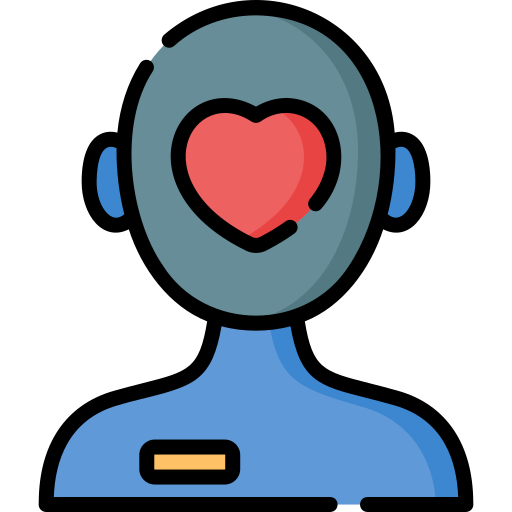}\hspace{0.5em}
\textbf{Believability (Human-likeness, Engagement)} examines whether the interaction conveys a sense of authenticity and naturalness that sustains user engagement from a human perspective.

\noindent \includegraphics[height=1.1em]{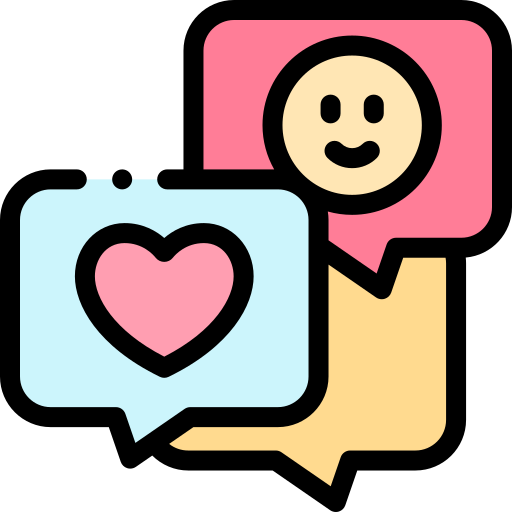}\hspace{0.5em}
\textbf{Conversational Quality \& Safety (Redundancy, Consistency, Safety)} assesses whether personalization is achieved without compromising coherence, stability, or ethical reliability, thereby ensuring a safe, consistent, and trustworthy interaction environment.

Detailed definitions of each evaluation dimension are provided in Appendix~\ref{app:eval_dimension}.

\section{Experimental Setup}
\label{sec: implementation_details}

\subsection{Human Study}
\label{sec: human study}

The Human Study is designed to collect basic user profiles and conduct human evaluations of human–AI interaction. Participants completed a demographic questionnaire, and suitable individuals were selected for the experiment. The detailed process is provided in Appendix \ref{app: human study}.

\paragraph{Participant Selection.}
Candidates completed the questionnaire described in Section~\ref{sec: User Profile Construction}. Eligible participants had prior experience with LLMs, a clearly defined personal issue to discuss, and stable psychological conditions. From over 600 submissions, 50 participants from diverse backgrounds were selected. They received training, reviewed sample dialogues, and studied evaluation guidelines to ensure consistency.

\paragraph{Human Interactive Evaluation.}
Each participant interacted with five models (Doubao-Pro, Qwen2.5-72B, GPT-4o, Claude-3.7-Sonnet, and DeepSeek-R1) in a blind, randomized order. After each session (minimum 10 turns), participants completed an evaluation questionnaire. Post-study interviews filtered out unserious participants, ensuring data quality. In total, 183 valid dialogues were collected, forming the \textbf{EmoHarbor Dataset}, used to assess alignment between automated metrics and human judgments.

\subsection{LLMs}
\label{sec: llm}

To ensure a comprehensive evaluation, this study employs a diverse set of LLMs, encompassing open-source, closed-source, and specialized models. The selected models are categorized as follows:

\paragraph{Open-Source Models.}
This category includes models from the Qwen family (Qwen-2.5~\cite{qwen}, Qwen-3~\cite{yang2025qwen3}, QwQ-32B~\cite{qwq}), the DeepSeek family (DeepSeek-R1~\cite{deepseekai2025deepseekr1}, DeepSeek-V3.1~\cite{deepseek-v3.1}), and GLM-4.5~\cite{glm4p5}\footnote{The DeepSeek models and GLM-4.5 were accessed via API due to their high computational requirements, rather than through local deployment.}.

\paragraph{Closed-Source Models.}
We also evaluate several state-of-the-art proprietary LLMs available through API services, including the Doubao family (Doubao-Seed-1.6~\cite{seed1.6}, Doubao-Pro), the Claude family (Claude-3.7-Sonnet and Claude-4-Sonnet)~\cite{claude4}, the Gemini family~\cite{gemini2.5}, and the GPT family (GPT-4o~\cite{gpt4o}, GPT-4~\cite{gpt-4}, GPT-5~\cite{gpt-5}, and o3-mini~\cite{o3}).

\paragraph{Specialized In-Domain Models.}
Finally, we incorporate models that have been fine-tuned specifically for mental health and emotional support applications: SoulChat~\cite{chen-etal-2023-soulchat}, PsyChat~\cite{qiu2024psychat}, and MindChat\footnote{\url{https://github.com/X-D-Lab/MindChat}}.

\subsection{Implementation Details}
\label{sec:implementation}

\paragraph{Experimental Environment.}
All experiments are conducted on 6 NVIDIA L40 GPUs. Our implementation is based on Python 3.12 and PyTorch 2.7.0, with inference accelerated using \texttt{vLLM}~\cite{kwon2023efficient}. 

\paragraph{Model Configurations.}
For model-specific configurations, GPT-4o is employed as both the User Thinker and User Talker agents, while Qwen3-235B serves as the User Evaluator agent. Temperature parameters are carefully chosen to align with each component's role: a low temperature of 0.1 for the User Thinker ensures focused and deterministic reasoning, whereas a higher temperature of 0.7 for the User Talker encourages diverse and natural responses. The User Evaluator operates at a temperature of 0.0 to guarantee consistent and reproducible assessments. All evaluated LLMs use a temperature of 0.7 during inference to maintain a balance between response diversity and coherence.

\paragraph{Simulation Configurations.}
Drawing on prior research in ESC \cite{ESConv}, the maximum number of User-Support interaction turns is set to 15. However, the User Agent is permitted to terminate the conversation prematurely by generating dialogue-ending signals, such as “\textit{Goodbye},” “\textit{Bye},” “\textit{That’s all},” or “\textit{I don’t want to continue}.”

\begin{table*}[ht]
\centering
\footnotesize
\setstretch{1.1}
\resizebox{\textwidth}{!}{%
\begin{tabular}{lcccccccccccc}
\toprule[1.2pt]
    \multicolumn{13}{l}{\makecell[l]{
    \textbb{PR}: Problem Resolution ~~~
    \textbb{MI}: Mood Improvement~~~
    \textbb{RA}: Response Appropriateness ~~~
    \textbb{AS}: Adaptive Strategies ~~~
    \\
    \textbb{EG}: Engagement ~~~  
    \textbb{HL}: Human-likeness ~~~
    \textbb{EP}: Empathetic ~~~
    \textbb{SF}: Safety ~~~
    \textbb{CS}: Consistency ~~~ 
    \textbb{RD}: Redundancy ~~~  
    }} \\
\midrule
\rowcolor[HTML]{D5F6F2} 
\multicolumn{1}{c}{\cellcolor[HTML]{D5F6F2}\textbf{Judge Model}} &
  \textbf{Profile} &
  \textbf{Internal State} &
  \textbf{\textbb{PR}} &
  \textbf{\textbb{MI}} &
  \textbf{\textbb{RA}} &
  \textbf{\textbb{AS}} &
  \textbf{\textbb{EG}} &
  \textbf{\textbb{HL}} &
  \textbf{\textbb{EP}} &
  \textbf{\textbb{SF}} &
  \textbf{\textbb{CS}} &
  \textbf{\textbb{RD}} \\ \midrule
 &
   &
   &
  0.35 &
  0.27 &
  0.18 &
  0.27 &
  0.36 &
  0.37 &
  0.21 &
  0.42 &
  0.41 &
  0.34 \\
 &
$\checkmark$ &
   &
  0.43 &
  0.46 &
  0.29 &
  0.29 &
  0.38 &
  0.42 &
  0.29 &
  0.41 &
  0.44 &
  0.39 \\
\multirow{-3}{*}{DeepSeek-R1} &
$\checkmark$ &
$\checkmark$ &
  \textbf{0.54} &
  \textbf{0.48} &
  \textbf{0.41} &
  \textbf{0.40} &
  \textbf{0.50} &
  \textbf{0.45} &
  \textbf{0.43} &
  \textbf{0.48} &
  \textbf{0.44} &
  \textbf{0.47} \\ \midrule
 &
   &
   &
  { 0.38} &
  { 0.42} &
  { 0.10} &
  { 0.18} &
  { 0.34} &
  { 0.37} &
  { 0.27} &
  { \textbf{0.54}} &
  { \textbf{0.41}} &
  { 0.29} \\
 &
$\checkmark$ &
   &
  { 0.41} &
  { 0.53} &
  { 0.20} &
  { 0.26} &
  { 0.32} &
  { 0.40} &
  { 0.26} &
  { 0.49} &
  { 0.40} &
  { 0.35} \\
\multirow{-3}{*}{Kimi-K2} &
$\checkmark$ &
$\checkmark$ &
  { \textbf{0.56}} &
  { \textbf{0.61}} &
  { \textbf{0.33}} &
  { \textbf{0.45}} &
  { \textbf{0.50}} &
  { \textbf{0.43}} &
  { \textbf{0.44}} &
  { 0.46} &
  { 0.40} &
  { \textbf{0.43}}\\ \midrule
 &
   &
   &
  0.20 &
  0.41 &
  0.36 &
  0.34 &
  0.22 &
  0.27 &
  0.35 &
  0.38 &
  0.37 &
  0.26 \\
 &
$\checkmark$ &
   &
  0.24 &
  0.47 &
  0.36 &
  0.36 &
  0.34 &
  0.25 &
  0.36 &
  0.37 &
  \textbf{0.43} &
  0.34 \\
\multirow{-3}{*}{GPT-4} &
$\checkmark$ &
$\checkmark$ &
  \textbf{0.42} &
  \textbf{0.57} &
  \textbf{0.41} &
  \textbf{0.45} &
  \textbf{0.41} &
  \textbf{0.29} &
  \textbf{0.33} &
  \textbf{0.40} &
  0.42 &
  \textbf{0.39} \\ \midrule
 &
   &
   &
  0.35 &
  0.43 &
  0.28 &
  0.22 &
  0.43 &
  0.35 &
  0.31 &
  0.43 &
  0.41 &
  0.29 \\
 &
$\checkmark$ &
   &
  0.49 &
  0.54 &
  0.32 &
  0.40 &
  \textbf{0.54} &
  0.39 &
  0.39 &
  \textbf{0.46} &
  0.40 &
  0.37 \\
\multirow{-3}{*}{Qwen3-235B} &
$\checkmark$ &
$\checkmark$ &
  \textbf{0.57} &
  \textbf{0.61} &
  \textbf{0.44} &
  \textbf{0.46} &
  \textbf{0.54} &
  \textbf{0.45} &
  \textbf{0.41} &
  \textbf{0.46} &
  \textbf{0.43} &
  \textbf{0.44} \\ \bottomrule[1.2pt]
\end{tabular}%
}
\vspace{-2mm}
\caption{Pearson correlation between model judgments and human assessments on EmoHarbor Dataset.}
\label{tab:judge_consistency}
\vspace{-3mm}
\end{table*}

\paragraph{Evaluation Configurations.}
Each dimension is rated on a 5-point Likert scale, with higher scores indicating better support. Detailed descriptions of the dimensions and the full evaluation protocol are provided in Appendix~\ref{app:eval_dimension}.

\section{Experimental Results}
\label{results}

In this section, we present experimental results to address the following key research questions:  

\noindent \textbf{Q1:} How reliable is the EmoHarbor Evaluation Framework? \\
\noindent \textbf{Q2:} How do existing models perform on the EmoHarbor Benchmark? \\ 
\noindent \textbf{Q3:} How do models adapt to user-specific needs in multi-turn interactions? 

\subsection{Empirical Validation of EmoHarbor Evaluation Framework}
\label{sec:empirical-validation}

\begin{figure}[th]
    \centering
    \vspace{-2mm}
    \includegraphics[width=\linewidth]{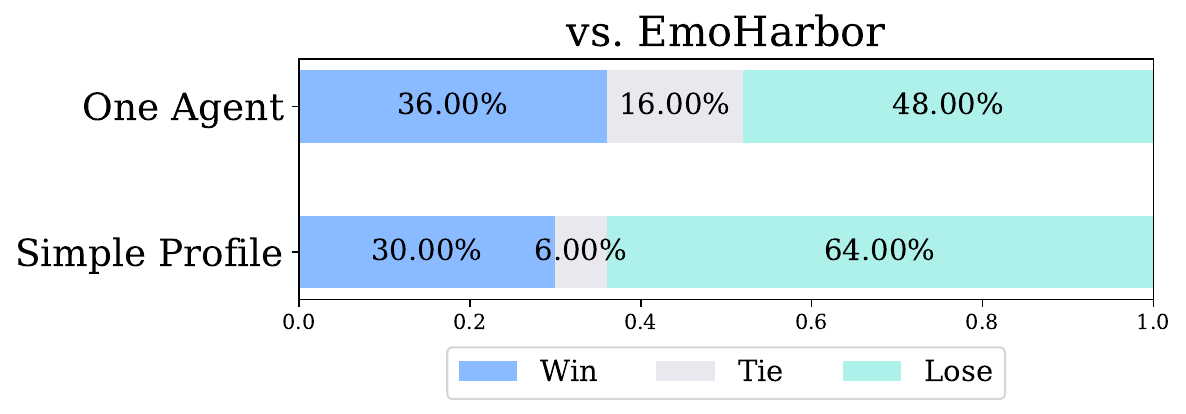}
    \vspace{-2mm}
    \caption{Pairwise human evaluation of User Simulator. `One Agent' lacks the User Thinker, generating responses directly by the User Talker. The `Simple Profile' uses only basic demographic and counseling attributes, excluding personality, preferences, and scenario scripts. {\color[HTML]{78e7dc}$\blacksquare$} indicates `EmoHarbor wins'.}  
    \vspace{-3mm}
    \label{fig:human_evaluation}
\end{figure}

\subsubsection{Human-Like Dialogue Generation.}

We conduct pairwise human evaluations to examine whether decomposing user simulation into multiple agents yields more human-like dialogues. The comparison includes three settings: (1) \textbf{One Agent}, in which a single model performs user simulation without explicit modeling of the user’s internal world; (2) \textbf{Simple Profile}, which conditions the simulator only on basic demographic and counseling-related attributes, without detailed user preference modeling or scenario scripts; and (3) our full Chain-of-Agent simulator, which incorporates Thinker and Talker modules operating over complete user profiles. Human judges are presented with pairs of dialogues and asked to select “A wins,” “Tie,” or “B wins,” with the presentation order randomized to mitigate positional bias. Evaluations are conducted on 50 randomly sampled dialogues.

The results in Figure~\ref{fig:human_evaluation} show that our Chain-of-Agent simulator consistently outperforms both baselines. Compared with One Agent, this demonstrates that modeling a user’s internal state produces responses that better reflect their personality and role. Against Simple Profile, our simulator achieves a 64\% win rate, indicating that incorporating richer, personalized features significantly improves the agent’s ability to engage in realistic role-playing.

\subsubsection{Alignment with Human Assessment.}
\label{sec:human-alignment}
We evaluate 4 candidate LLM judges—DeepSeek-R1, Kimi-K2, GPT-4, and Qwen3-235B—on the EmoHarbor Dataset under the following strategies:
(1) \textbf{Standard Judgment.} The evaluator rates the emotional-support quality of each dialogue based solely on the conversation text. This strategy represents the conventional setup, in which evaluation is limited to the observable dialogue without additional context.
(2) \textbf{User-Aware Judgment.} The evaluator considers both the conversation text and the corresponding user profile. Incorporating user-specific information makes the assessment more personalized and context-sensitive.
(3) \textbf{User-Internal-State-Aware Judgment (ours).} Beyond the conversation text and user profile, we simulate the user’s internal state at each turn using a user thinker agent, given the preceding dialogue. These simulated states approximate the user’s inferred thoughts and emotions and are used to inform the evaluation. The original dialogue content remains unchanged; the internal states serve solely as auxiliary context to improve assessment fidelity.

Table~\ref{tab:judge_consistency} shows Pearson correlations with human ratings. The overall alignment across all models is moderate, with values around 0.4–0.5. While these values may seem modest, they are consistent with the inherent subjectivity of personalized emotional support evaluation. In this context, a correlation in this range indicates that the LLM judges are reasonably capturing human judgment and can serve as a practical and usable evaluation signal. Importantly, incorporating user profiles and turn-level user states further improves this alignment, particularly for highly subjective and personalized dimensions such as PR, MI, RA, and AS.

\subsubsection{Benchmark Discrimination Capability.}
\label{sec:discrimination-capability}

\begin{table}[th]
\centering
\footnotesize
\setstretch{1.2}
\begin{adjustbox}{width=\linewidth}
\begin{tabular}{cccc}
\toprule[1.2pt]
\textbf{MSR} & \textbf{MAC} & \textbf{ANOVA}            & \textbf{\begin{tabular}[c]{@{}c@{}}Pairwise\\  Discriminability\end{tabular}} \\ \midrule
0.745        & 0.427        & F=112 (p\textless{}0.001) & 0.87                                                                          \\ \bottomrule[1.2pt]
\end{tabular}
\end{adjustbox}
\vspace{-1mm}
\caption{Overall benchmark discrimination performance. Model Separation Ratio (MSR) measures the strength of inter-model performance differences relative to user-level rating noise. Model Agreement Coefficient (MAC) quantifies the consistency of user judgments when comparing models.}
\label{tab:discrimination-metrics}
\end{table}

We evaluate the discriminative power of the EmoHarbor evaluation framework, which examines how effectively the User-as-a-Judge paradigm can distinguish performance differences among ESC models.  
The detailed computation of each metric is provided in Appendix~\ref{app: metircs}. 
As summarized in Table~\ref{tab:discrimination-metrics}, the benchmark achieves an MSR (Model Separation Ratio) of 0.745, indicating that inter-model differences are substantially larger than user-level rating noise.  
The MAC (Model Agreement Coefficient) of 0.427 reflects moderate-to-strong consistency among user judgments when comparing models.    
These results are further corroborated by a significant one-way ANOVA result ($F = 112, p < 0.001$) and a high pairwise discriminability score (0.87), showing that users can reliably differentiate between model performances.  
Collectively, these findings demonstrate that EmoHarbor possesses strong discriminative capability under the User-as-a-Judge paradigm.

\subsection{Benchmark Results}
\label{sec:benchmark results}

\begin{table*}[th]
\centering
\footnotesize
\setstretch{1.05}
\resizebox{\textwidth}{!}{%
\begin{tabular}{lcccccccccccc}
\toprule[1.2pt]
    \multicolumn{13}{l}{\makecell[l]{
    \textbb{PR}: Problem Resolution ~~~
    \textbb{MI}: Mood Improvement~~~
    \textbb{RA}: Response Appropriateness ~~~
    \textbb{AS}: Adaptive Strategies ~~~ \\
    \textbb{EG}: Engagement ~~~ 
    \textbb{HL}: Human-likeness ~~~ 
    \textbb{EP}: Empathetic ~~~
    \textbb{SF}: Safety ~~~
    \textbb{CS}: Consistency ~~~ 
    \textbb{RD}: Redundancy ~~~  
    }} \\
\midrule
\multicolumn{1}{l}{
\textbf{Models}} &
  \multicolumn{1}{l}{
  \textbf{Reasoning}} &
  \textbf{\textbb{PR}} &
  \textbf{\textbb{MI}} &
  \textbf{\textbb{RA}} &
  \textbf{\textbb{AS}} &
  \textbf{\textbb{EG}} &
  \textbf{\textbb{HL}} &
  \textbf{\textbb{EP}} &
  \textbf{\textbb{SF}} &
  \textbf{\textbb{CS}} &
  \textbf{\textbb{RD}} &
  \textbf{Avg.} \\ \midrule
\rowcolor[HTML]{EFEFEF} 
\multicolumn{13}{l}{\cellcolor[HTML]{EFEFEF}\textit{\textbf{Open-Source}}}                                                                        \\ 
\midrule
Qwen2.5-7B-Instruct &
   &
  \cellcolor[HTML]{E3FBF8}1.80 &
  \cellcolor[HTML]{A2F1E6}1.59 &
  \cellcolor[HTML]{64E8D6}1.39 &
  \cellcolor[HTML]{7AEBDC}1.46 &
  \cellcolor[HTML]{ABF3E9}1.62 &
  \cellcolor[HTML]{E4ECFE}2.57 &
  \cellcolor[HTML]{F4F7FF}2.18 &
  \cellcolor[HTML]{BFD1FD}3.52 &
  \cellcolor[HTML]{BACEFD}3.64 &
  \cellcolor[HTML]{FBFCFF}1.98 &
  2.18 \\
Qwen2.5-32B-Instruct &
   &
  \cellcolor[HTML]{E9FCF9}2.07 &
  \cellcolor[HTML]{B5F4EB}1.93 &
  \cellcolor[HTML]{64E8D6}1.71 &
  \cellcolor[HTML]{98F0E4}1.85 &
  \cellcolor[HTML]{AEF3EA}1.91 &
  \cellcolor[HTML]{E3EBFE}2.83 &
  \cellcolor[HTML]{F1F5FF}2.47 &
  \cellcolor[HTML]{BCCFFD}3.79 &
  \cellcolor[HTML]{BACEFD}3.84 &
  \cellcolor[HTML]{FDFDFF}2.19 &
  2.46 \\
Qwen2.5-72B-Instruct &
   &
  \cellcolor[HTML]{C5F6F0}2.12 &
  \cellcolor[HTML]{A4F2E7}2.04 &
  \cellcolor[HTML]{64E8D6}1.88 &
  \cellcolor[HTML]{A0F1E6}2.03 &
  \cellcolor[HTML]{E1FBF7}2.19 &
  \cellcolor[HTML]{DBE5FE}3.18 &
  \cellcolor[HTML]{F1F5FF}2.63 &
  \cellcolor[HTML]{BED1FD}3.91 &
  \cellcolor[HTML]{BACEFD}4.02 &
  \cellcolor[HTML]{FCFDFF}2.34 &
  2.63 \\
Qwen3-32B &
$\checkmark$ &
  \cellcolor[HTML]{64E8D6}2.23 &
  \cellcolor[HTML]{8FEEE1}2.28 &
  \cellcolor[HTML]{BAF5ED}2.33 &
  \cellcolor[HTML]{6DE9D8}2.24 &
  \cellcolor[HTML]{FCFDFF}2.48 &
  \cellcolor[HTML]{D3E0FE}3.45 &
  \cellcolor[HTML]{EFF3FF}2.80 &
  \cellcolor[HTML]{BACEFD}4.05 &
  \cellcolor[HTML]{BED1FD}3.95 &
  \cellcolor[HTML]{C3F6EF}2.34 &
  2.82 \\
Qwen3-235B &
$\checkmark$ &
  \cellcolor[HTML]{64E8D6}\textbf{3.67} &
  \cellcolor[HTML]{E1FBF7}\textbf{3.86} &
  \cellcolor[HTML]{9FF1E6}3.76 &
  \cellcolor[HTML]{CEF8F2}\textbf{3.83} &
  \cellcolor[HTML]{FCFDFF}\textbf{3.95} &
  \cellcolor[HTML]{D2DFFE}\textbf{4.50} &
  \cellcolor[HTML]{E1EAFE}\textbf{4.30} &
  \cellcolor[HTML]{BACEFD}\textbf{4.82} &
  \cellcolor[HTML]{C0D2FD}\textbf{4.74} &
  \cellcolor[HTML]{D4F9F4}\textbf{3.84} &
  \textbf{4.13} \\
QwQ-32B &
$\checkmark$ &
  \cellcolor[HTML]{93EFE2}3.59 &
  \cellcolor[HTML]{64E8D6}3.53 &
  \cellcolor[HTML]{FCFDFF}3.76 &
  \cellcolor[HTML]{AAF2E8}3.62 &
  \cellcolor[HTML]{E8FCF9}3.70 &
  \cellcolor[HTML]{D3E0FE}4.24 &
  \cellcolor[HTML]{C9F7F1}3.66 &
  \cellcolor[HTML]{BBCFFD}4.52 &
  \cellcolor[HTML]{BACEFD}4.53 &
  \cellcolor[HTML]{FCFDFF}3.76 &
  3.89 \\
DeepSeek-V3.1 &
$\checkmark$ &
  \cellcolor[HTML]{72EADA}3.11 &
  \cellcolor[HTML]{64E8D6}3.09 &
  \cellcolor[HTML]{79EBDB}3.12 &
  \cellcolor[HTML]{86EDDF}3.14 &
  \cellcolor[HTML]{E7FBF9}3.28 &
  \cellcolor[HTML]{D6E2FE}4.06 &
  \cellcolor[HTML]{F4F7FF}3.52 &
  \cellcolor[HTML]{BACEFD}4.58 &
  \cellcolor[HTML]{BDD0FD}4.52 &
  \cellcolor[HTML]{FDFEFF}3.35 &
  3.58 \\
DeepSeek-R1 &
$\checkmark$ &
  \cellcolor[HTML]{7CECDC}3.49 &
  \cellcolor[HTML]{64E8D6}3.45 &
  \cellcolor[HTML]{EEF3FF}\textbf{3.94} &
  \cellcolor[HTML]{ADF3E9}3.57 &
  \cellcolor[HTML]{E4FBF8}3.66 &
  \cellcolor[HTML]{CEDCFE}4.38 &
  \cellcolor[HTML]{FCFDFF}3.75 &
  \cellcolor[HTML]{C3D4FD}4.53 &
  \cellcolor[HTML]{BACEFD}4.65 &
  \cellcolor[HTML]{ADF3E9}3.57 &
  3.90 \\
GLM-4.5 &
$\checkmark$ &
  \cellcolor[HTML]{94EFE3}2.88 &
  \cellcolor[HTML]{88EDE0}2.85 &
  \cellcolor[HTML]{64E8D6}2.76 &
  \cellcolor[HTML]{9CF0E5}2.90 &
  \cellcolor[HTML]{D5F9F4}3.04 &
  \cellcolor[HTML]{DEE7FE}3.77 &
  \cellcolor[HTML]{F6F8FF}3.32 &
  \cellcolor[HTML]{BACEFD}4.44 &
  \cellcolor[HTML]{BCD0FD}4.40 &
  \cellcolor[HTML]{F9FBFF}3.25 &
  3.36 \\
\midrule
\rowcolor[HTML]{EFEFEF} 
\multicolumn{13}{l}{\cellcolor[HTML]{EFEFEF}\textit{\textbf{Closed-Source}}}                                                                       \\ \midrule
Doubao-Seed-1.6 &
$\checkmark$ &
  \cellcolor[HTML]{64E8D6}3.63 &
  \cellcolor[HTML]{9BF0E4}3.69 &
  \cellcolor[HTML]{DBFAF5}3.76 &
  \cellcolor[HTML]{92EFE2}\textbf{3.68} &
  \cellcolor[HTML]{FCFDFF}3.84 &
  \cellcolor[HTML]{CDDBFE}4.46 &
  \cellcolor[HTML]{E8EFFE}4.10 &
  \cellcolor[HTML]{BBCFFD}4.70 &
  \cellcolor[HTML]{BACEFD}4.71 &
  \cellcolor[HTML]{6DE9D8}3.64 &
  4.02 \\
Doubao-Pro-32k &
  \multicolumn{1}{l}{} &
  \cellcolor[HTML]{92EFE2}2.21 &
  \cellcolor[HTML]{92EFE2}2.21 &
  \cellcolor[HTML]{64E8D6}2.08 &
  \cellcolor[HTML]{7DECDD}2.15 &
  \cellcolor[HTML]{D6F9F4}2.40 &
  \cellcolor[HTML]{D3DFFE}3.56 &
  \cellcolor[HTML]{F5F8FF}2.74 &
  \cellcolor[HTML]{BFD2FD}4.02 &
  \cellcolor[HTML]{BACEFD}4.14 &
  \cellcolor[HTML]{FAFCFF}2.63 &
  2.81 \\
Claude-4-Sonnet &
$\checkmark$ &
  \cellcolor[HTML]{64E8D6}3.41 &
  \cellcolor[HTML]{64E8D6}3.41 &
  \cellcolor[HTML]{F7F9FF}3.76 &
  \cellcolor[HTML]{B8F4EC}3.54 &
  \cellcolor[HTML]{84EDDF}3.46 &
  \cellcolor[HTML]{D4E1FE}4.23 &
  \cellcolor[HTML]{E8EEFE}3.97 &
  \cellcolor[HTML]{BACEFD}4.59 &
  \cellcolor[HTML]{BED1FD}4.53 &
  \cellcolor[HTML]{B8F4EC}3.54 &
  3.84 \\
Claude-3.7-Sonnet &
   &
  \cellcolor[HTML]{72EADA}3.16 &
  \cellcolor[HTML]{64E8D6}3.14 &
  \cellcolor[HTML]{D2F8F3}3.30 &
  \cellcolor[HTML]{9BF0E5}3.22 &
  \cellcolor[HTML]{E0FAF7}3.32 &
  \cellcolor[HTML]{D8E3FE}4.06 &
  \cellcolor[HTML]{EDF2FE}3.69 &
  \cellcolor[HTML]{BCCFFD}4.56 &
  \cellcolor[HTML]{BACEFD}4.59 &
  \cellcolor[HTML]{FCFDFF}3.41 &
  3.65 \\
Gemini-2.5-Pro &
$\checkmark$ &
  \cellcolor[HTML]{64E8D6}3.42 &
  \cellcolor[HTML]{C5F6F0}\textbf{3.76} &
  \cellcolor[HTML]{9AF0E4}3.61 &
  \cellcolor[HTML]{ABF3E9}3.67 &
  \cellcolor[HTML]{FFFFFF}\textbf{3.97} &
  \cellcolor[HTML]{CEDCFE}\textbf{4.60} &
  \cellcolor[HTML]{DAE4FE}\textbf{4.45} &
  \cellcolor[HTML]{BBCFFD}\textbf{4.85} &
  \cellcolor[HTML]{BACEFD}\textbf{4.86} &
  \cellcolor[HTML]{FEFFFF}\textbf{3.96} &
  \textbf{4.12} \\
GPT-4o-2024-11-20 &
  \multicolumn{1}{l}{} &
  \cellcolor[HTML]{B0F3EA}2.98 &
  \cellcolor[HTML]{DDFAF6}3.09 &
  \cellcolor[HTML]{64E8D6}2.79 &
  \cellcolor[HTML]{C5F6F0}3.03 &
  \cellcolor[HTML]{F9FEFD}3.16 &
  \cellcolor[HTML]{D9E4FE}3.96 &
  \cellcolor[HTML]{E6EDFE}3.70 &
  \cellcolor[HTML]{BACEFD}4.60 &
  \cellcolor[HTML]{BED1FD}4.51 &
  \cellcolor[HTML]{FEFEFF}3.19 &
  3.50 \\
GPT-5-2025-08-07 &
$\checkmark$ &
  \cellcolor[HTML]{F6FEFD}\textbf{3.64} &
  \cellcolor[HTML]{64E8D6}3.31 &
  \cellcolor[HTML]{F2F6FF}\textbf{3.80} &
  \cellcolor[HTML]{FFFFFF}3.66 &
  \cellcolor[HTML]{8CEEE1}3.40 &
  \cellcolor[HTML]{E9EFFE}3.90 &
  \cellcolor[HTML]{FFFFFF}3.66 &
  \cellcolor[HTML]{C1D3FD}4.33 &
  \cellcolor[HTML]{BACEFD}4.41 &
  \cellcolor[HTML]{F6FEFD}3.64 &
  3.77 \\
o3-mini &
$\checkmark$ &
  \cellcolor[HTML]{ABF3E9}2.36 &
  \cellcolor[HTML]{64E8D6}2.25 &
  \cellcolor[HTML]{B8F4EC}2.38 &
  \cellcolor[HTML]{91EFE2}2.32 &
  \cellcolor[HTML]{B2F4EB}2.37 &
  \cellcolor[HTML]{D4E1FE}3.65 &
  \cellcolor[HTML]{E7EEFE}3.15 &
  \cellcolor[HTML]{BACEFD}4.36 &
  \cellcolor[HTML]{BDD0FD}4.28 &
  \cellcolor[HTML]{FBFCFF}2.60 &
  2.97 \\
 \midrule
\rowcolor[HTML]{EFEFEF} 
\multicolumn{13}{l}{\cellcolor[HTML]{EFEFEF}\textit{\textbf{Specialized In-Domain}}}                                                                          \\ \midrule
SoulChat2.0-Qwen2-7B &
  \multicolumn{1}{l}{} &
  \cellcolor[HTML]{D8F9F5}1.35 &
  \cellcolor[HTML]{AEF3EA}1.23 &
  \cellcolor[HTML]{64E8D6}1.02 &
  \cellcolor[HTML]{84EDDE}1.11 &
  \cellcolor[HTML]{8BEEE0}1.13 &
  \cellcolor[HTML]{E5EDFE}2.05 &
  \cellcolor[HTML]{FAFCFF}1.57 &
  \cellcolor[HTML]{BDD0FD}2.98 &
  \cellcolor[HTML]{BACEFD}3.04 &
  \cellcolor[HTML]{F1F5FF}1.77 &
  1.73 \\
PsyChat-Qwen2.5-7B &
  \multicolumn{1}{l}{} &
  \cellcolor[HTML]{B2F4EB}2.16 &
  \cellcolor[HTML]{77EBDB}2.10 &
  \cellcolor[HTML]{C5F6F0}2.18 &
  \cellcolor[HTML]{64E8D6}2.08 &
  \cellcolor[HTML]{CFF8F2}2.19 &
  \cellcolor[HTML]{D5E1FE}\textbf{3.41} &
  \cellcolor[HTML]{E7EEFE}2.92 &
  \cellcolor[HTML]{BED1FD}4.06 &
  \cellcolor[HTML]{BACEFD}\textbf{4.16} &
  \cellcolor[HTML]{FDFEFF}2.29 &
  2.75 \\
MindChat-Qwen-7B-v2 &
  \multicolumn{1}{l}{} &
  \cellcolor[HTML]{CFF8F2}\textbf{2.61} &
  \cellcolor[HTML]{E5FBF8}\textbf{2.67} &
  \cellcolor[HTML]{64E8D6}\textbf{2.32} &
  \cellcolor[HTML]{9FF1E6}\textbf{2.48} &
  \cellcolor[HTML]{DAFAF5}\textbf{2.64} &
  \cellcolor[HTML]{E5EDFE}3.31 &
  \cellcolor[HTML]{F1F5FF}\textbf{3.04} &
  \cellcolor[HTML]{BACEFD}\textbf{4.25} &
  \cellcolor[HTML]{C0D2FD}4.12 &
  \cellcolor[HTML]{FCFDFF}\textbf{2.81} &
  \textbf{3.02} \\
\bottomrule[1.2pt]
\end{tabular}%
}
\vspace{-3mm}
\caption{Evaluation results of LLMs on\textbf{ \ourbench}. All scores are on a 5-point Likert scale.  For each section, the best performance is highlighted in \textbf{bold}. For each model, dimensions with strong performance are highlighted in \colorbox{myblue}{``Blue''}, while weaker performance is highlighted in \colorbox{mygreen}{``Green''}. Darker shades indicate more extreme performance.}
\vspace{-3mm}
\label{tab:benchmark_results}
\end{table*}
Table~\ref {tab:benchmark_results} presents the evaluation results, highlighting the following key observations:

\paragraph{Existing LLMs are still far from expert-level performance on personalized \task.}  
We evaluate a diverse set of LLMs on \ourbench, including the Qwen, DeepSeek, Claude, GPT, Gemini, and Doubao families. Among closed-source systems, Gemini-2.5-Pro achieves the best overall performance, with a peak score of 4.12. Other models perform worse, with most failing to exceed a score of 4. Among open-source systems, Qwen3-235B performs best, achieving an average score of 4.13 and competitive results compared to closed-source models. This strong performance may be partly attributed to Chinese being its primary training and research language. Besides, when comparing reasoning-oriented models (RLMs) to non-RLMs, we observe that RLMs consistently perform better across both open-source and closed-source families. Notably, most RLMs achieve scores above 3, indicating a clear advantage in handling personalized reasoning-intensive tasks.  

\paragraph{Specialized in-domain LLMs also struggle with \task.}  
Previous studies have shown that many conversational LLMs are heavily optimized for empathetic response generation, often reporting promising results on benchmarks such as ESConv when evaluated with BLEU or ROUGE metrics. However, these improvements do not generalize well to personalized \task. For instance, SoulChat2.0 \cite{chen-etal-2023-soulchat} achieves an average score of only 1.73. This underperformance is likely due to overfitting on empathetic dialogue datasets, which limits the model’s ability to adapt responses based on individual user characteristics.  

\paragraph{LLMs show solid basic conversational skills but fail to provide effective emotional support.}  
Our analysis reveals that almost all models perform better on dimensions such as human-likeness, consistency, empathy, and safety, compared to dimensions like problem resolution, mood improvement, engagement, and personalization. Engagement scores, in particular, remain low, suggesting that conversations often feel ineffective and may even have negative side effects. This highlights important directions for future improvements in emotional support and user-centered adaptation.

\subsection{Analysis of Multi-turn Performance}
\label{sec:conversational-dynamics}
\begin{figure}[t]
\vspace{-3mm}
    \centering
    \includegraphics[width=\linewidth]{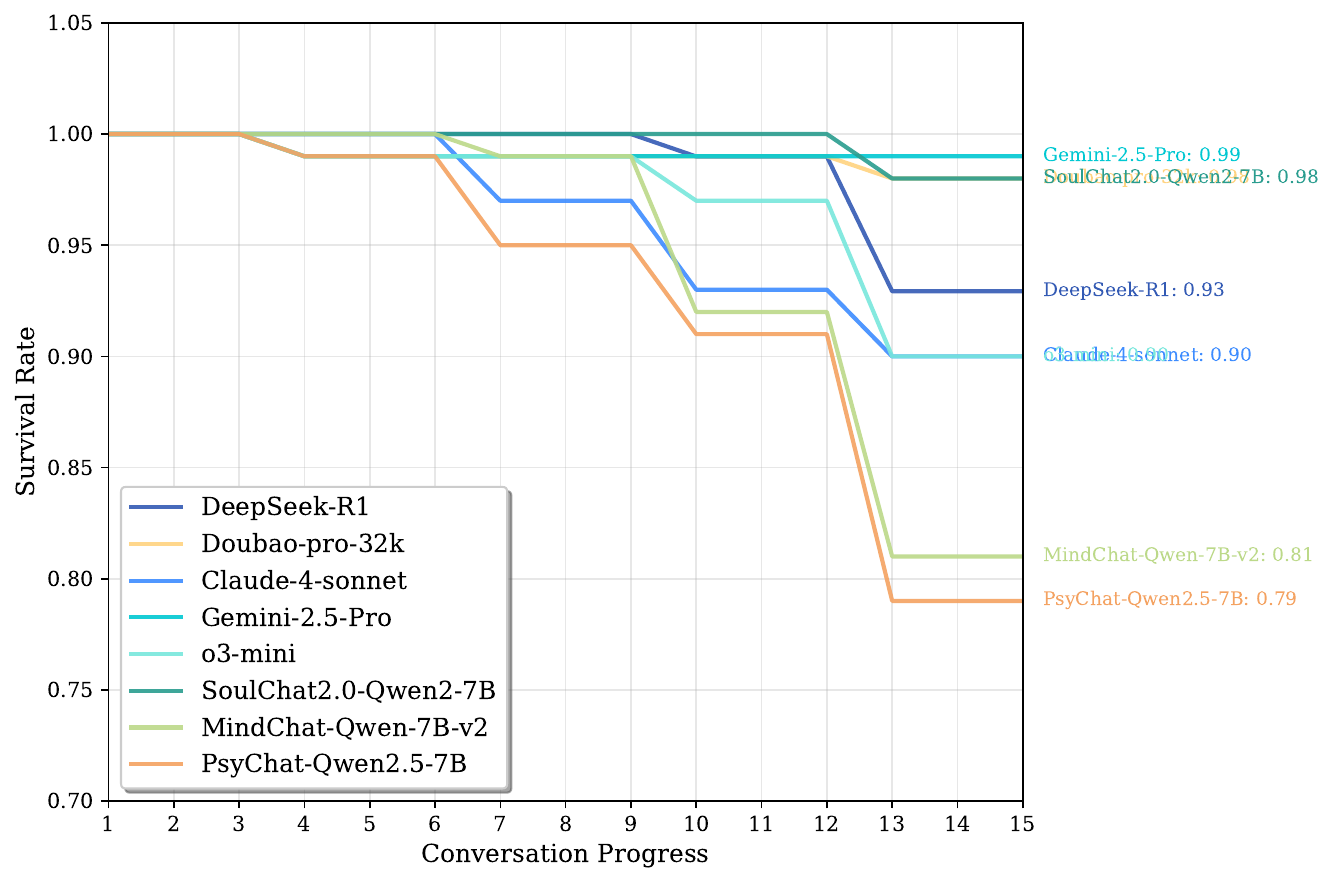}
    \vspace{-2mm}
    \caption{Survival rates of the models as the conversation progresses.}
    \vspace{-2mm}
    \label{fig:survival_curve}
\end{figure}

\begin{figure}[t]
    \centering
    \begin{subfigure}[b]{0.45\linewidth}
        \centering
        \includegraphics[width=\linewidth]{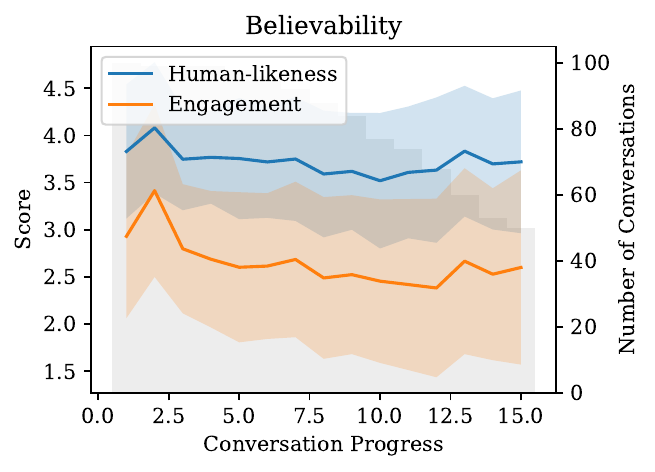}
    \end{subfigure}
    \hfill
    \begin{subfigure}[b]{0.45\linewidth}
        \centering
        \includegraphics[width=\linewidth]{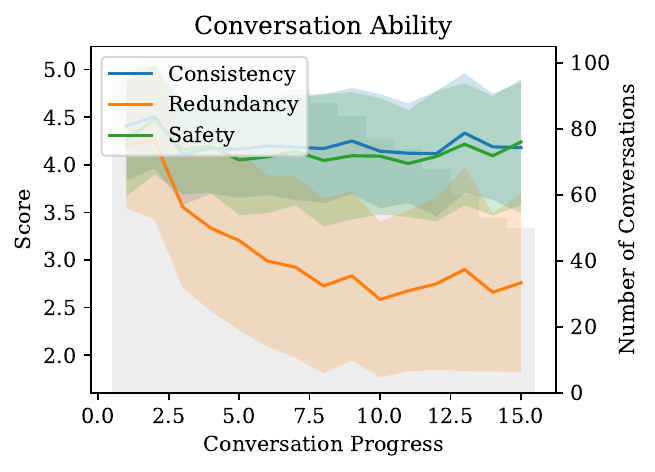}
    \end{subfigure}
    \begin{subfigure}[b]{0.45\linewidth}
        \centering
        \includegraphics[width=\linewidth]{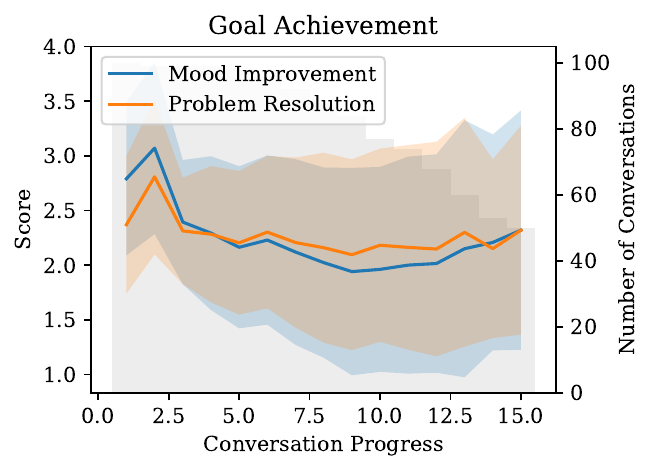}
    \end{subfigure}
    \hfill
    \begin{subfigure}[b]{0.45\linewidth}
        \centering
        \includegraphics[width=\linewidth]{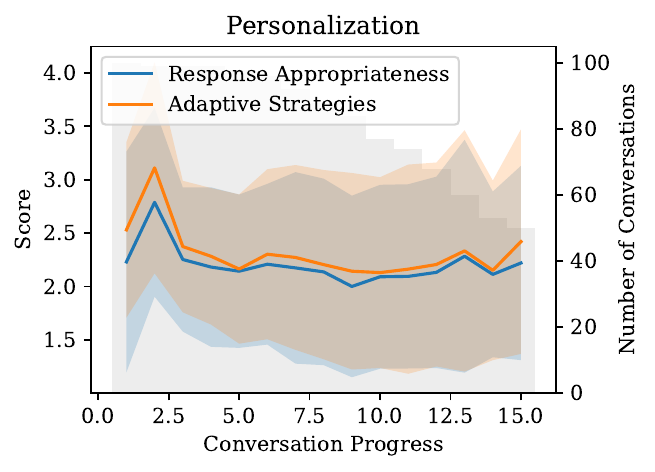}
    \end{subfigure}
    \caption{Multi-turn dialogue evaluation of Doubao-Pro. Lines represent the average performance at each dialogue turn, while the gray background indicates the number of conversations that reached the corresponding turn.}
    \label{fig:multi_turn_doubao}
\end{figure}

As mentioned in Section~\ref{sec:implementation}, we set the maximum dialogue length to 15 turns, slightly below the average 17–18 turns observed in real ESC conversations \cite{ESConv}. However, we observed that in many cases the \textit{User Agent} actively ended the conversation. The primary reason for early termination is that the ESC system fails to provide effective emotional support, leading to disengagement. We regard such early terminations as indicative of model failure.

Figure~\ref{fig:survival_curve} presents the survival curves of dialogue sessions for different models. From the curves, we can see that models with better overall performance also tend to sustain conversations for longer periods. Figure~\ref{fig:multi_turn_doubao} details how Doubao-Pro’s performance in each dimension changes as the dialogue progresses. Among the completed dialogues, the model’s overall performance remains relatively stable, though all metrics exhibit a slight downward trend as the conversation progresses. This suggests potential weaknesses in maintaining quality over extended periods of interaction. Notably, the redundancy score declines markedly with increasing dialogue turns, implying that as conversations become longer, the model tends to produce repetitive or formulaic responses, leading to less effective empathetic engagement.

\section{Related Work}
With the advancement of LLMs, personalized ES agents \cite{pal, ye2025generi, suh2025sense7, SocialSim, jiang2025artificial, jiang2025knowmerespondme} have attracted growing research interest. A key challenge persists: how to effectively evaluate the quality of emotional support.

\paragraph{Traditional Evaluation.} Early ESC evaluation \citep{ESConv, AugESC, ExTES, ESCoT, serveforemo} relied on automatic metrics like BLEU \cite{bleu}, ROUGE \cite{rouge}, and BERTScore \cite{BERTScore}, which assess token overlap or embedding similarity with references. These metrics often \textit{fail to capture ESC’s diversity and nuance}. Human evaluation, though the gold standard, is \textit{slow, costly, and subjective, yielding low inter-annotator agreement and poor reproducibility} \cite{ESC-Judge}.

\paragraph{Specialist Judge Evaluation.} Fine-tuned judge models, such as CharacterEval \cite{tu-etal-2024-charactereval} and CharacterBench \cite{characterbench}, use annotated data as specialist evaluators. While more scalable than human evaluation, they have key limitations: (1) \textbf{Static dialogue}—they rely on pre-collected dialogue logs, failing to capture real-time interactivity or evolving conversational context; (2) \textbf{Context bias}—dialogue histories are not self-generated by the evaluated model, and are influenced by the model’s in-context learning, leading to bias and inadequate assessment of multi-turn dialogue capabilities \cite{cpo}. \citet{ESC-Eval, ESC-Judge} partially address these issues with a user–supporter simulation framework. Still, their evaluation focuses excessively on language fluency and empathetic expression while neglecting users’ personalized needs.

\paragraph{LLM-as-a-Judge Evaluation.} Recent studies use LLMs as scalable judges, offering alternatives to human annotation and static benchmarks \cite{zheng2023judgingllmasajudgemtbenchchatbot, gu2025surveyllmasajudge, batcheval, kazi2024largelanguagemodelsuseragents}. Sotopia \cite{Sotopia} assesses emotional intelligence via role-playing simulations, while ESC-Judge \cite{ESC-Judge} and CharacterArena \cite{cpo} adopt a user simulator to generate dialogues for pairwise comparison \cite{chatbotarena}. Despite these advancements, they fall short in capturing the user-centric, context-sensitive, and psychologically grounded nature of emotional support evaluation. A truly effective evaluation framework should shift toward personalized, interaction-aware, and subjectively grounded assessment strategies that reflect users’ real emotional experiences.

\section{Conclusion}
This paper proposes EmoHarbor, a simple yet effective evaluation framework that addresses the challenge of assessing personalized emotional support conversations. EmoHarbor leverages a user-as-a-judge paradigm through a chain-of-agent architecture, moving beyond conventional homogeneous expert judgments. Experiments on 20 advanced LLMs show that while current LLMs excel at generic empathy, they struggle to provide user-tailored support. This work presents a novel and efficient pathway to developing more nuanced and user-aware emotional support systems.

\section*{Limitations}
This study presents a novel evaluation framework for personalized emotional support conversations, grounded in a user-as-a-judge paradigm. The proposed framework offers new directions for advancing the development of more nuanced and user-aware emotional support systems. Nonetheless, several limitations merit further consideration.
\textit{Firstly, }although the user simulation encompasses a variety of user profiles, it is constructed upon predefined structures and may not fully capture the complexity and unpredictability of real human behavior. 
\textit{Secondly, }the human consistency evaluation may be influenced by participants’ understanding of the evaluation task and their familiarity with LLMs, potentially introducing systematic biases that are difficult to eliminate.

\section*{Ethical Considerations}

This research utilized publicly available models, including Deepseek \cite{deepseekai2025deepseekr1}, Qwen \cite{qwen2.5}, GLM \cite{glm4p5}, Doubao \cite{doubao}, Claude \cite{claude-3-5}, Gemini \cite{gemini2.5}, and GPT \cite{gpt-4}, as well as toolkits such as \texttt{vLLM} \cite{kwon2023efficient}.

The benchmark datasets used in our evaluation were synthetically generated using GPT-4o and are scheduled for public release upon acceptance. The profiles used in this study were manually verified and filtered; however, we cannot guarantee that the content generated by user agents and support agents is entirely harmless due to the inherent unpredictability of LLMs. The primary language of focus in this work is Chinese. This study is intended solely for research purposes.

We adhered to strict ethical guidelines in our human study. Fifty participants from diverse backgrounds were recruited. Before beginning the evaluation, participants received a clear and thorough explanation of the study's objectives, potential risks, and the evaluation process. To ensure fair compensation and respect for their time, participants were paid 50 CNY per hour, a rate exceeding the prevailing local labor standard. All participant data will be kept confidential and will not be disclosed without explicit consent.

LLMs were employed to assist in coding, writing, and polishing the manuscript. Importantly, the LLMs were not involved in the ideation, research methodology, or experimental design. All research concepts, ideas, and analyses were developed and conducted solely by the authors.


\bibliography{custom}
\clearpage
\onecolumn
\appendix
\section*{Appendix}

\startcontents[sections]
\printcontents[sections]{l}{1}{\setcounter{tocdepth}{2}}

\clearpage
\twocolumn

\section{EmoHarbor Dataset}
\label{app: human study}
We developed the EmoHarbor Dataset through controlled human studies designed to capture human–AI dialogues. Each entry records an authentic conversation between a participant and an AI model, together with the participant’s profile and their subjective evaluation of the model’s responses. These evaluations reflect the user’s individual perspectives and emotional context, providing a rich foundation for studying personalized human–AI interactions. The human–AI interaction interface used for data collection is described in Appendix \ref{app: human evaluation interface}.

\paragraph{Participants Selection.}
Prospective participants were required to complete the initial questionnaire described in Section \ref{sec: User Profile Construction}. To be eligible, participants needed prior experience with LLMs, a clearly defined personal issue to discuss during the experiment, and a stable psychological condition. In total, we received more than 600 questionnaire submissions. Based on the completeness and quality of these responses, we selected 50 participants from diverse backgrounds for the human–AI interaction evaluation. The selected participants attended a training session to familiarize themselves with the experimental setup. They reviewed example dialogues and detailed evaluation guidelines to ensure consistent and meaningful ratings.

\paragraph{Conversational Dataset Collection.}
Each participant interacted with five models from different families (Doubao-Pro, Qwen2.5-72B, GPT-4o, Claude-3.7-Sonnet, and DeepSeek-R1), presented in a blind, randomized order. Each interaction consisted of at least ten conversational turns, allowing participants to explore topics of personal relevance in depth. After each session, participants completed a structured evaluation questionnaire to express their subjective judgments of the model and its alignment with their emotional needs.
To ensure data quality, post-experiment interviews were conducted to identify and exclude participants who did not engage seriously with the tasks. Ultimately, we obtained 183 valid human–AI dialogue instances, which together constitute the EmoHarbor Dataset. This dataset enables systematic analysis of the alignment between automated evaluation metrics and authentic, user-centered human judgments.

\section{User Profile Construction}
\label{app: user profile construction}
\paragraph{Chinese User Profile Construction.}
Chinese user profiles are constructed from seed information collected during the data acquisition of the EmoHarbor dataset. Specifically, participants provided basic background descriptions and brief counseling-related problem statements through preliminary questionnaires. Following the user profile definition in Section~\ref{sec: User Profile Construction}, we systematically instantiate each component of the profile as follows.
(1) \textit{Demographic attributes} are rewritten to remove identifiable details while preserving essential contextual grounding.
(2) \textit{Preference-related attributes} are expanded to enhance individual variability, including personality traits, habits, and speech style.
(3) \textit{Counseling-related attributes} are concretized by elaborating on the event background, emotional state, and user goals based on the original responses.
(4) Finally, a \textit{scenario script} is constructed to specify plausible emotional and behavioral reactions to different types of counselor feedback.
This structured construction process ensures that simulated users are both diverse and internally consistent, thereby mitigating behavioral homogenization in role-playing.

\begin{figure}[t]
    \centering
    \includegraphics[width=\linewidth]{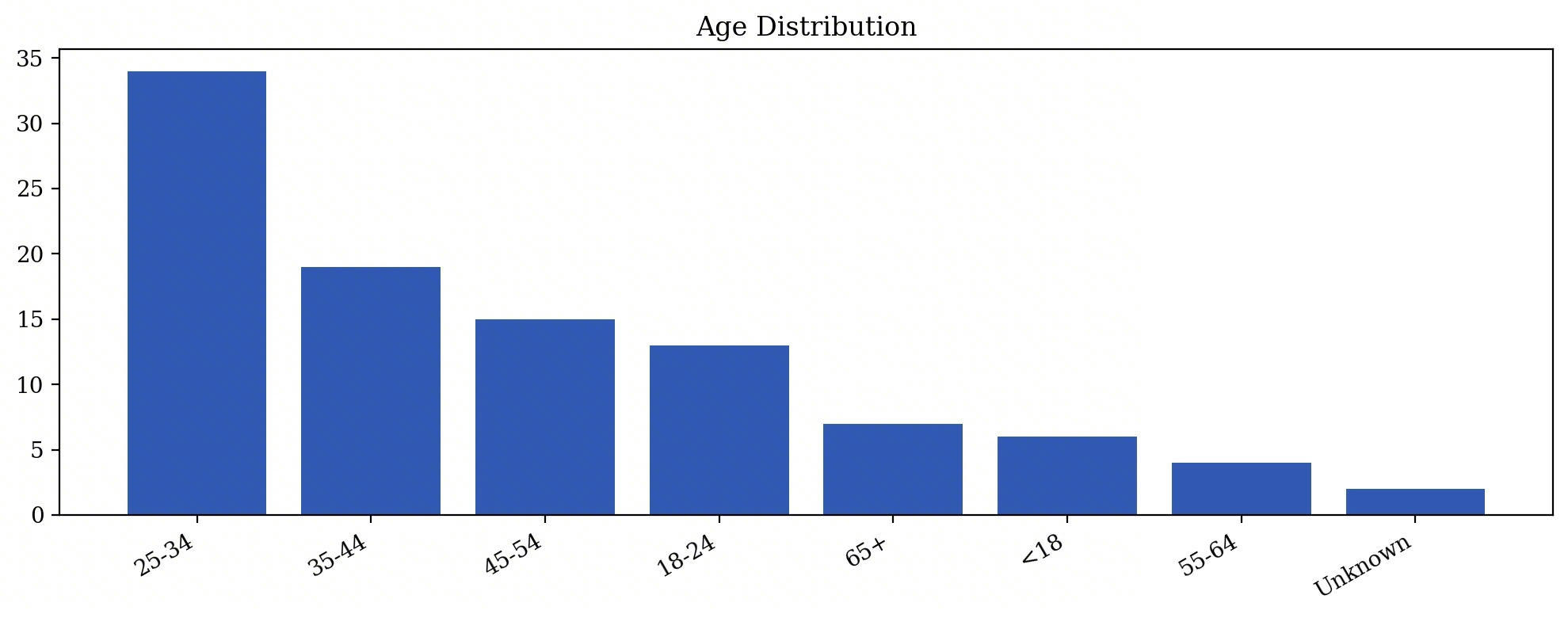}\\
    \includegraphics[width=\linewidth]{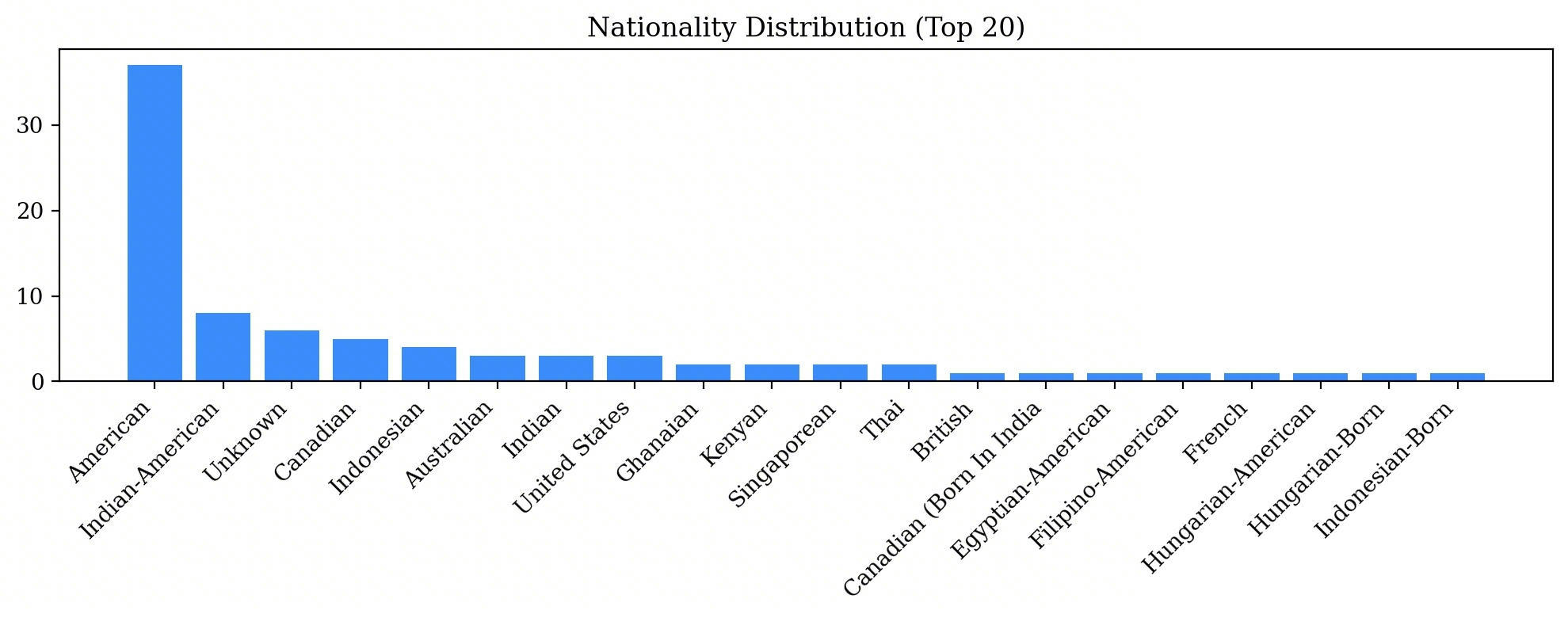}\\
    \includegraphics[width=\linewidth]{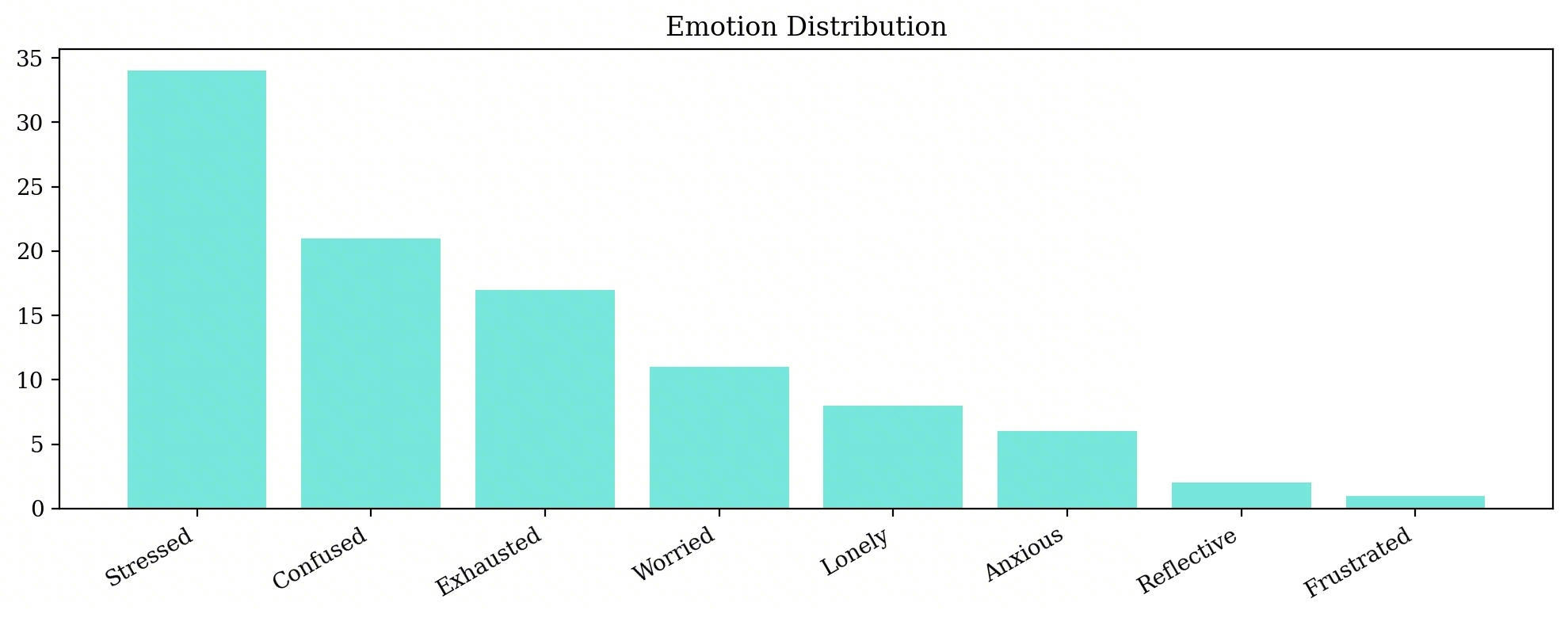}\\
    \includegraphics[width=\linewidth]{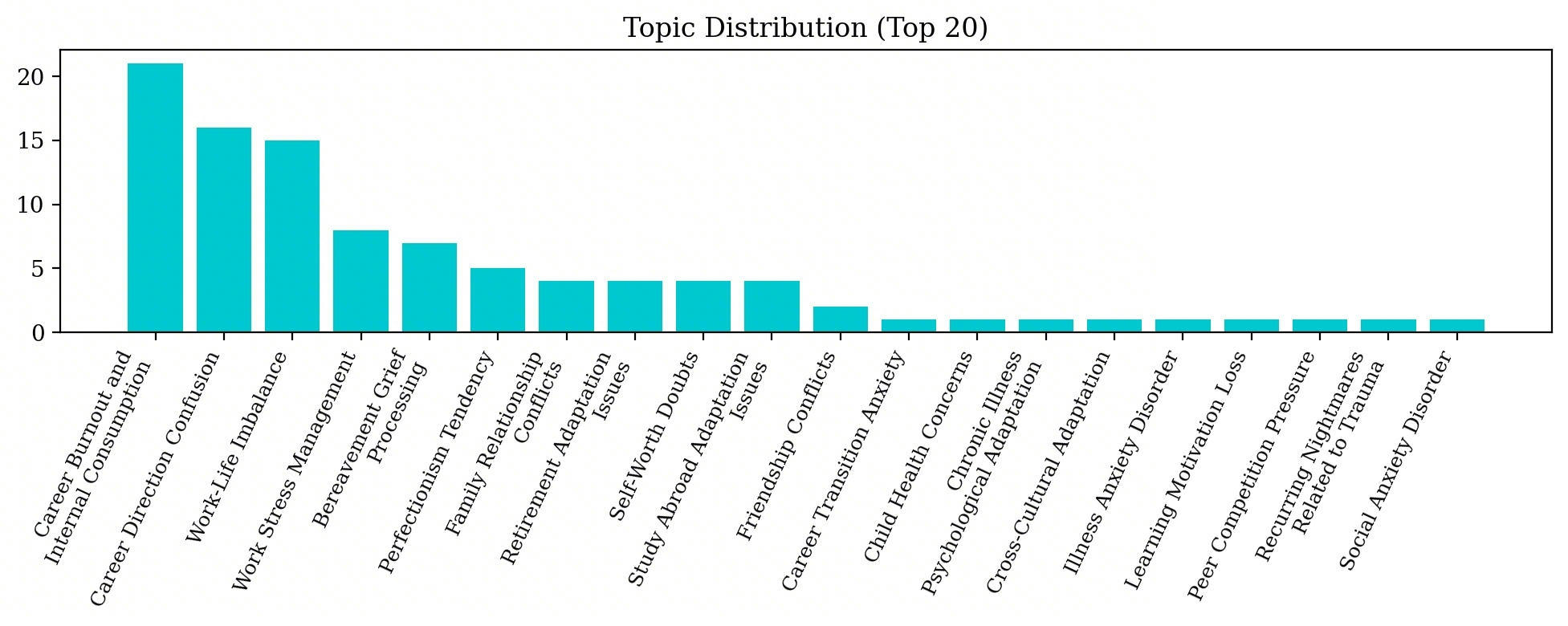}
    \caption{English User Profile Distribution.}
    \label{fig:english user profile distribution}
\end{figure}

\paragraph{English User Profile Construction.}
Unlike Chinese profiles, English user profiles are initialized from an existing profile set proposed by \citet{personamemv2}, which provides rich demographic and preference-related attributes. Building on the counseling problem categories in ESConv, we further construct counseling-related attributes for each profile by specifying the corresponding emotional context, problem background, and the user's seek goals. Scenario scripts are then authored following the same procedure used for the Chinese user profiles, defining plausible emotional and behavioral responses to different types of counselor feedback. All constructed profiles are first automatically validated using a large language model and subsequently manually inspected via random sampling for quality control. We ultimately retain 100 high-quality English user profiles for use in our experiments. Figure~\ref{fig:english user profile distribution} summarizes the distributions of age, nationality, emotional states, and problem topics.

\section{EmoHarbor Evaluation Workflow}
\label{app: algorithm}

We present the workflow for evaluating emotional-support dialogue systems using EmoHarbor in Algorithm~\ref{alg:EmoHarbor}. The evaluation simulates multi-turn interactions between a user and the system under test, while maintaining both the user’s internal state and conversation history. At the end of the dialogue, the user model produces structured evaluation scores across multiple dimensions, reflecting the system’s performance in providing personalized and emotionally attuned support.

\begin{algorithm}[ht]
\small
\caption{Workflow of EmoHarbor Evaluation}
\label{alg:EmoHarbor}
\begin{algorithmic}[1]
    \Require User Profile $\mathcal{P_U}$, Supporter System Under Test $\mathcal{S}$, Max turns $T$  
    \Ensure Evaluation scores $E^{1:K}$ on $K$ dimensions
    \State Initialize Supporter memory $H_s \leftarrow \emptyset$, User memory $H_u \leftarrow \emptyset$
    \State Initialize user internal state $IS \leftarrow \text{InitialState}(\mathcal{P_U})$
    \For{$t \gets 1, T$}
        \State $R_t \leftarrow \mathcal{S}(H_s)$ \Comment{Get system response}
        \State $IS_t \leftarrow \text{UserThinker}(H_u, \mathcal{P_U}, R_t)$ \Comment{Update internal state}
        \State $U_t \leftarrow \text{UserTalker}(H_u, IS, \mathcal{P_U}, R_t)$ 
        \Comment{Generate user utterance}
        \State $H_s \leftarrow H_s \cup \{(R_t), (U_t)\}$ \Comment{Update Supporter memory}   
        \State $H_u \leftarrow H_u \cup \{(R_t), (IS_t), (U_t)\}$ \Comment{Update User memory}  
    \EndFor
    \State $E^{1:K} \leftarrow \text{UserEvaluator}(H_u, \mathcal{P_U})$ \Comment{Final evaluation}
    \State \Return $E^{1:K}$
\end{algorithmic}
\end{algorithm}

\section{LLMs}
\label{app: llm}
 
\paragraph{SoulChat.} SoulChat\cite{chen-etal-2023-soulchat} is a Chinese dialogue model designed to enhance empathy, active listening, and comforting abilities. It is instruction-tuned on SoulChatCorpus, a multi-turn empathetic dialogue dataset, to strengthen its emotional support capabilities. The dataset contains 2,300,248 psychological counseling questions across 12 topics.

\paragraph{PsyChat.} PsyChat\cite{qiu2024psychat} is a client-centric dialogue system for mental health support. It consists of five core modules: client behavior recognition, counselor strategy selection, input packing, response generation, and response selection. This modular design enables adaptive and personalized interactions that align with the user’s emotional state.

\paragraph{MindChat.} MindChat\footnote{\url{https://github.com/X-D-Lab/MindChat}}
 is a Chinese dialogue model designed for real-world mental health support scenarios. It is trained on approximately one million high-quality multi-turn psychological counseling dialogues automatically constructed through a rule-based data generation process. The dataset covers various domains, including work, family, study, daily life, social interactions, and safety. Owing to its unique data construction methodology, MindChat is capable of engaging users in more empathetic and guiding conversations.


\section{Discriminative Ability Metrics}
\label{app: metircs}
To assess whether the user-as-a-judge evaluation framework can reliably distinguish performance differences among emotional support conversation systems, we introduce a set of quantitative metrics that capture the benchmark’s discriminative ability. Specifically, we measure how consistently the User Agent (hereafter referred to simply as the user) perceives differences between models, and how pronounced those differences are relative to user-level rating noise.

Let $U$ denote the number of users, $M$ the number of models, and $r_{u,m}$ the rating given by user $u$ to model $m$. The overall mean rating, denoted by $\bar{r}$, is computed as:
\begin{equation}
\bar{r} = \frac{1}{UM} \sum_{u=1}^{U} \sum_{m=1}^{M} r_{u,m}.
\end{equation}

For a specific model $m$, its average rating $\bar{r}_m$ is computed over all users:
\begin{equation} 
\bar{r}_m = \frac{1}{U} \sum_{u=1}^{U} r_{u,m} 
\end{equation}

\subsection{Model Separation Ratio (MSR).}
To quantify the disparity in model performance relative to user rating consistency, we use the Model Separation Ratio (MSR). This metric is derived from the between-model variance and the within-model variance.

The between-model variance $\sigma_{\text{between}}^2$ measures the dispersion of individual model performances around the grand mean. It quantifies how much, on average, the performance of each model deviates from the overall average performance.
\begin{equation}
\sigma_{\text{between}}^2 = \frac{1}{M} \sum_{m=1}^{M} (\bar{r}_m - \bar{r})^2
\end{equation}

The within-model variance, $\sigma_{\text{within}}^2$, measures the average dispersion of individual user ratings around each model's own mean. It reflects the consistency in user ratings for a given model, averaged across all models.
\begin{equation}
\sigma_{\text{within}}^2 = \frac{1}{M} \sum_{m=1}^{M} \frac{1}{U-1} \sum_{u=1}^{U} (r_{u,m} - \bar{r}_m)^2
\end{equation}

The MSR is then defined as the ratio of the between-model variance to the within-model variance. 
\begin{equation}
    \text{MSR} = \frac{\sigma_{\text{between}}^2}{\sigma_{\text{within}}^2}
\end{equation}
A higher MSR indicates that the differences in performance between models are large compared to the variation in user opinions for each model, suggesting that the models are more easily distinguishable.

\begin{figure*}[th]
    \centering
    \begin{subfigure}[b]{0.24\textwidth}
        \centering
        \includegraphics[width=\linewidth]{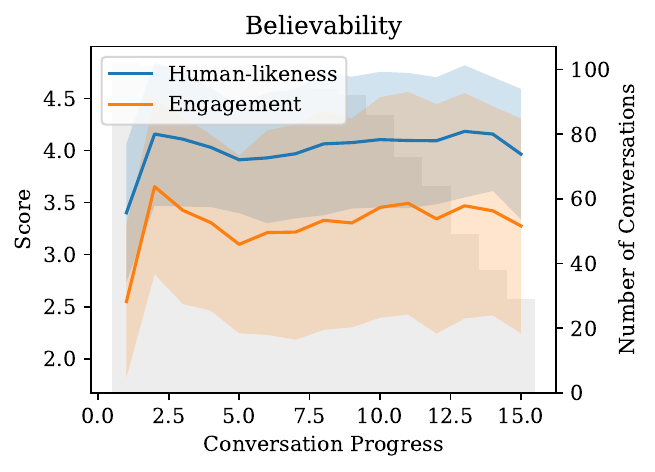}
    \end{subfigure}
    \hfill
    \begin{subfigure}[b]{0.24\textwidth}
        \centering
        \includegraphics[width=\linewidth]{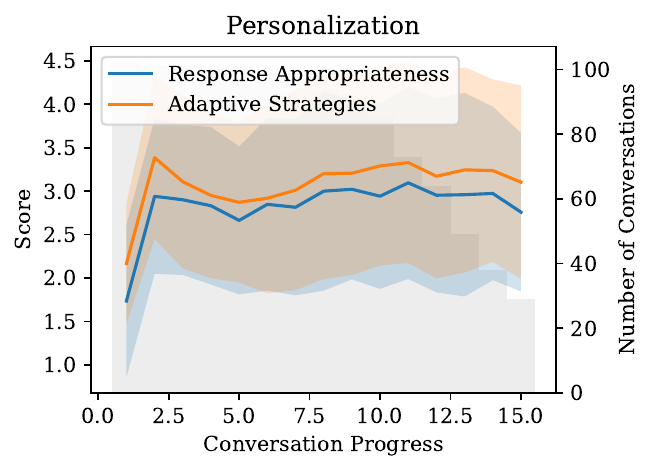}
    \end{subfigure}
    \hfill
    \begin{subfigure}[b]{0.24\textwidth}
        \centering
        \includegraphics[width=\linewidth]{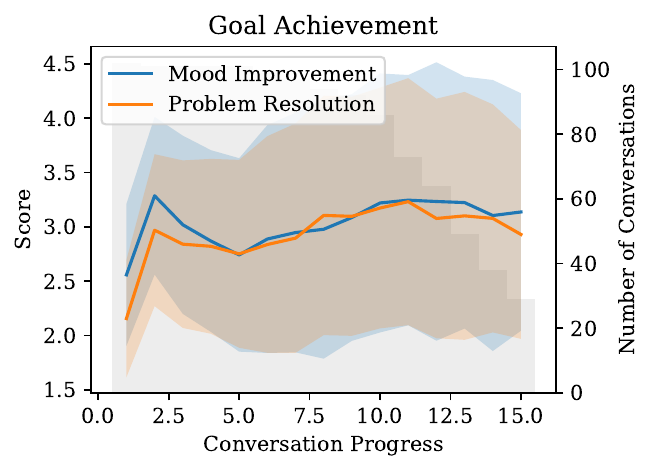}
    \end{subfigure}
    \hfill
    \begin{subfigure}[b]{0.24\textwidth}
        \centering
        \includegraphics[width=\linewidth]{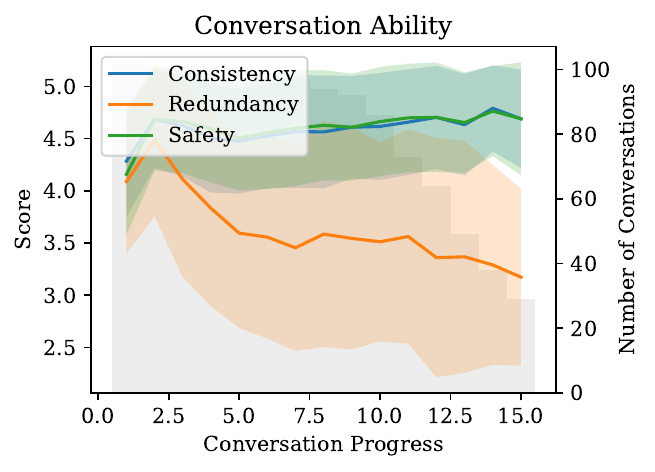}
    \end{subfigure}
    \caption{Multi-turn Dialogue Evaluation Experiment on GPT-4o.}
    \label{fig: multi-turn_4o_app}
    \hfill
    \begin{subfigure}[b]{0.24\textwidth}
        \centering
        \includegraphics[width=\linewidth]{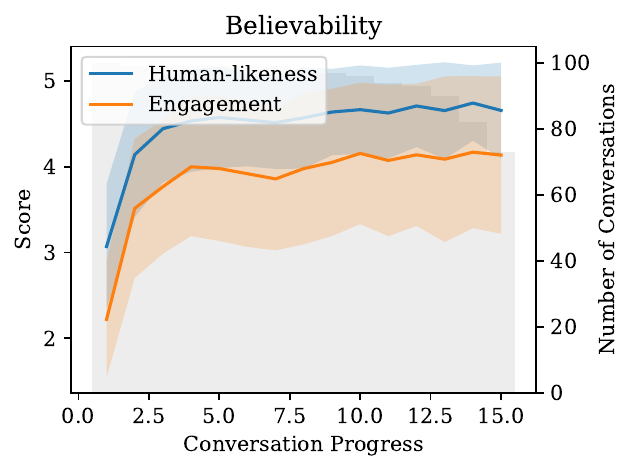}
    \end{subfigure}
    \hfill
    \begin{subfigure}[b]{0.24\textwidth}
        \centering
        \includegraphics[width=\linewidth]{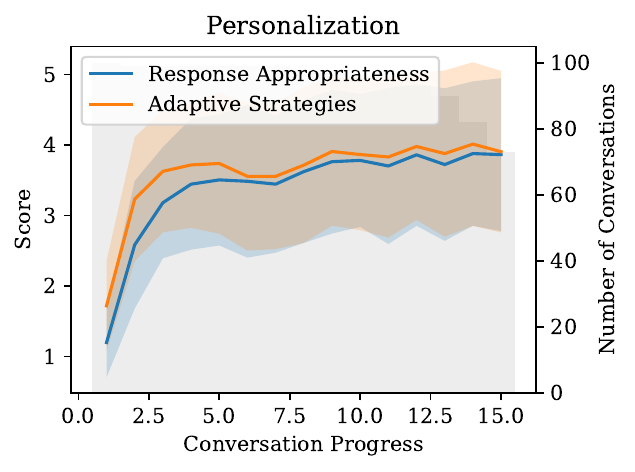}
    \end{subfigure}
    \hfill
    \begin{subfigure}[b]{0.24\textwidth}
        \centering
        \includegraphics[width=\linewidth]{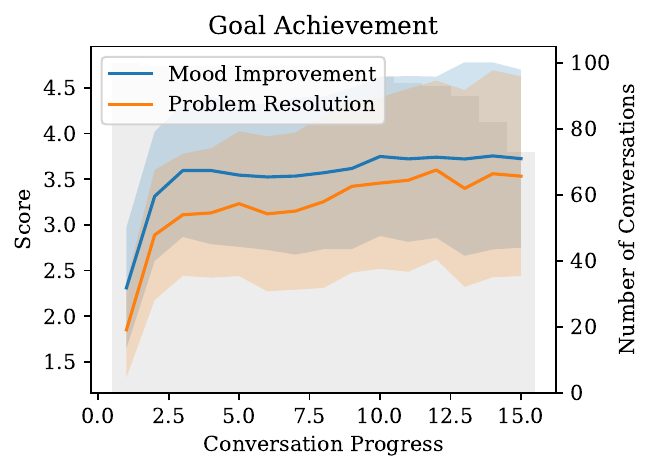}
    \end{subfigure}
    \hfill
    \begin{subfigure}[b]{0.24\textwidth}
        \centering
        \includegraphics[width=\linewidth]{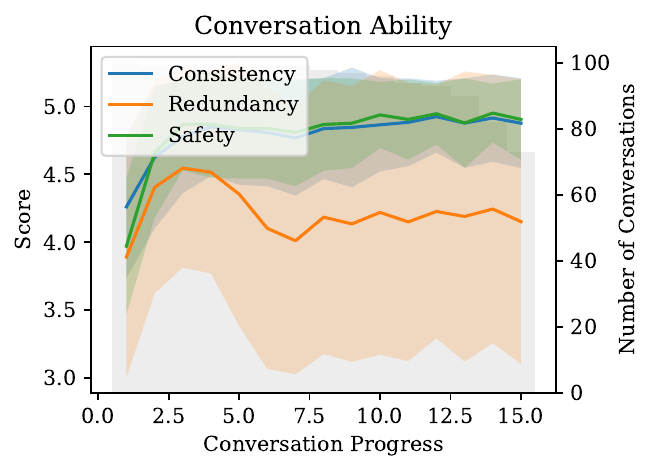}
    \end{subfigure}
    \caption{Multi-turn Dialogue Evaluation Experiment on Gemini-2.5-Pro.}
    \label{fig: multi-turn_gemini}
\end{figure*}

\subsection{Model Agreement Coefficient (MAC).}
The Model Agreement Coefficient (MAC) evaluates the degree of consensus among users when ranking models. It represents the proportion of the total rating variance attributable to systematic differences between models—rather than random disagreement across individual user judgments.
\begin{equation}
\text{MAC} = \frac{\sigma_{\text{between}}^2}{\sigma_{\text{between}}^2 + \sigma_{\text{within}}^2}.
\end{equation}
A MAC value close to 1 indicates strong inter-user agreement on the relative quality of models, implying that users consistently perceive model performance differences in emotional support dialogues. Conversely, a lower MAC suggests that subjective variation dominates, signaling weaker consensus.

\subsection{One-way ANOVA F-statistic}
To statistically verify whether performance differences among models are significant, we apply a one-way ANOVA with the model as the grouping factor. The resulting F-statistic tests whether model means differ beyond what could be explained by user-level variability:
\begin{equation}
F = \frac{\sigma_{\text{between}}^2 / (M-1)}{\sigma_{\text{within}}^2 / (M(U-1))}.
\end{equation}
A large F-value (with $p < 0.05$) indicates that at least one model’s mean rating significantly differs from others, confirming that users can reliably distinguish models’ emotional support quality.

\subsection{Pairwise Discriminability Proportion}
Finally, to capture the granularity of model distinctions, we compute the Pairwise Discriminability Proportion.  
For all pairs of models $(i,j)$, we count the number of pairs with statistically significant rating differences (after multiple-comparison correction), and compute:
\begin{equation}
P = \frac{\#\text{significant}}{\binom{M}{2}}.
\end{equation}
A high $P$ value reflects that users can consistently recognize pairwise differences in conversational or emotional support quality across models.

\section{Additional Experimental Results}
\label{app:experiments}

\subsection{English Benchmark Results}
\label{app:additional_benchmark_results}

In Section~\ref{sec:benchmark results}, we report benchmark results for the Chinese setting. Here, we present corresponding evaluations in the English setting, with results summarized in Table~\ref{tab:benchmark_results_en}.

Overall, the English results exhibit trends highly consistent with those observed in the Chinese benchmarks. RLMs consistently outperform non-RLMs, and in-domain training provides additional performance gains. Across models, performance is relatively weaker on \textit{Problem Resolution}, \textit{Mood Improvement}, \textit{Engagement}, and \textit{Redundancy}. This indicates that while current LLMs can generate empathetic responses at the turn level, they remain limited in addressing personalized user needs and maintaining non-redundant, engaging behavior over long conversations. Taken together, these findings further underscore that achieving personalized, long-horizon emotional companionship remains a challenging open problem.

\subsection{Additional Multi-turn Analysis Results}
\label{app: addtional multi-turn results}
As discussed in \cref{sec:conversational-dynamics}, we provide additional results on how the performance of GPT-4o and Gemini-2.5-Pro changes across different dimensions as the dialogue progresses. The results in Figures \ref{fig: multi-turn_4o_app} and \ref{fig: multi-turn_gemini} show that: (1) All models tend to exhibit decreasing scores for redundancy as the conversation continues. This indicates that in longer dialogues, models are prone to repeating patterns and producing redundant content. (2) Gemini-2.5-Pro performs relatively better across all dimensions. We observe that it maintains a higher retention rate throughout multi-turn dialogues. Moreover, its scores in certain dimensions even show an increasing trend as the conversation progresses. This suggests that Gemini-2.5-Pro is able to provide more user-relevant content over multiple turns, effectively engaging users and encouraging continued interaction.

\subsection{Cost Analysis}
\label{app:cost_analysis}

\begin{table}[th]
\centering
\footnotesize
\setstretch{1.2}
\begin{tabular}{cccc}
\toprule[1.2pt]
\textbf{Method} & \textbf{Time} & \textbf{Input Tokens} &  \textbf{Output Tokens} \\
\midrule
EmoHarbor & 188.8s & 92K & 2K \\
\bottomrule[1.2pt]
\end{tabular}
\caption{Average computational cost per dialogue in the User-as-a-Judge evaluation pipeline. The analyzed Supporter model is GPT-4o.}
\label{tab:cost_analysis}
\end{table}

The proposed User-as-a-Judge evaluation framework relies on multi-agent coordination and multi-turn user simulation, which inevitably incurs additional computational overhead. To enhance transparency and support informed adoption, we report detailed runtime and token-level cost statistics. Table~\ref{tab:cost_analysis} presents the average computational cost per evaluated dialogue under the full EmoHarbor setting. On average, evaluating a single dialogue requires 188.8 seconds, consuming approximately $92\mathrm{K}$ input tokens and $2\mathrm{K}$ output tokens. Most of the computational cost stems from iterative user simulation and reflective evaluation stages, which are critical for modeling user-level psychological dynamics. To mitigate evaluation cost while maintaining reasoning fidelity, we adopt \textbf{Qwen3-235B} as the User Evaluator model, striking a favorable balance between inference efficiency and reasoning capability.

\begin{table*}[t]
\centering
\footnotesize
\setstretch{1.05}
\resizebox{\textwidth}{!}{%
\begin{tabular}{lcccccccccccc}
\toprule[1.2pt]
    \multicolumn{13}{l}{\makecell[l]{
    \textbb{PR}: Problem Resolution ~~~
    \textbb{MI}: Mood Improvement~~~
    \textbb{RA}: Response Appropriateness ~~~
    \textbb{AS}: Adaptive Strategies ~~~ \\
    \textbb{EG}: Engagement ~~~ 
    \textbb{HL}: Human-likeness ~~~ 
    \textbb{EP}: Empathetic ~~~
    \textbb{SF}: Safety ~~~
    \textbb{CS}: Consistency ~~~ 
    \textbb{RD}: Redundancy ~~~  
    }} \\
\midrule
\multicolumn{1}{l}{
\textbf{Models}} &
  \multicolumn{1}{l}{
  \textbf{Reasoning}} &
  \textbf{\textbb{PR}} &
  \textbf{\textbb{MI}} &
  \textbf{\textbb{RA}} &
  \textbf{\textbb{AS}} &
  \textbf{\textbb{EG}} &
  \textbf{\textbb{HL}} &
  \textbf{\textbb{EP}} &
  \textbf{\textbb{SF}} &
  \textbf{\textbb{CS}} &
  \textbf{\textbb{RD}} &
  \textbf{Avg.} \\ \midrule
\multicolumn{13}{l}{\cellcolor[HTML]{EFEFEF}\textit{\textbf{Open-Source}}} \\ \midrule
Llama-3-8B-Instruct &
   &
  \cellcolor[HTML]{ACF3E9}2.40 &
  \cellcolor[HTML]{64E8D6}2.00 &
  \cellcolor[HTML]{F9FBFF}3.00 &
  \cellcolor[HTML]{E6FBF8}2.72 &
  \cellcolor[HTML]{9CF0E5}2.31 &
  \cellcolor[HTML]{D7E3FE}3.76 &
  \cellcolor[HTML]{EFF4FF}3.22 &
  \cellcolor[HTML]{BACEFD}4.42 &
  \cellcolor[HTML]{CDDCFE}3.99 &
  \cellcolor[HTML]{B9F5EC}2.47 &
  3.04 \\
Llama-3.1-8B-Instruct &
   &
  \cellcolor[HTML]{90EFE2}2.52 &
  \cellcolor[HTML]{64E8D6}2.27 &
  \cellcolor[HTML]{F1F5FF}3.40 &
  \cellcolor[HTML]{D3F8F3}2.90 &
  \cellcolor[HTML]{A7F2E8}2.65 &
  \cellcolor[HTML]{D6E2FE}3.89 &
  \cellcolor[HTML]{EBF1FE}3.52 &
  \cellcolor[HTML]{BACEFD}4.41 &
  \cellcolor[HTML]{C9D9FD}4.14 &
  \cellcolor[HTML]{B9F5EC}2.75 &
  3.25 \\
Qwen2.5-7B-Instruct &
   &
  \cellcolor[HTML]{EAFCFA}1.78 &
  \cellcolor[HTML]{76EBDB}1.27 &
  \cellcolor[HTML]{FAFBFF}2.03 &
  \cellcolor[HTML]{D1F8F3}1.67 &
  \cellcolor[HTML]{64E8D6}1.19 &
  \cellcolor[HTML]{EDF2FE}2.45 &
  \cellcolor[HTML]{FCFDFF}1.96 &
  \cellcolor[HTML]{BACEFD}4.04 &
  \cellcolor[HTML]{D7E3FE}3.12 &
  \cellcolor[HTML]{C8F7F1}1.63 &
  2.11 \\
Qwen2.5-32B-Instruct &
   &
  \cellcolor[HTML]{C0F6EE}2.36 &
  \cellcolor[HTML]{64E8D6}1.81 &
  \cellcolor[HTML]{F5F8FF}2.95 &
  \cellcolor[HTML]{DBFAF5}2.52 &
  \cellcolor[HTML]{A4F1E7}2.19 &
  \cellcolor[HTML]{D4E1FE}3.65 &
  \cellcolor[HTML]{F1F5FF}3.04 &
  \cellcolor[HTML]{BACEFD}4.22 &
  \cellcolor[HTML]{C8D8FD}3.91 &
  \cellcolor[HTML]{BBF5ED}2.33 &
  2.90 \\
Qwen3-8B &
 $\checkmark$&
  \cellcolor[HTML]{95EFE3}2.31 &
  \cellcolor[HTML]{64E8D6}2.03 &
  \cellcolor[HTML]{F3F6FF}3.15 &
  \cellcolor[HTML]{D5F9F4}2.67 &
  \cellcolor[HTML]{97F0E4}2.32 &
  \cellcolor[HTML]{D3E0FE}3.79 &
  \cellcolor[HTML]{F0F4FF}3.21 &
  \cellcolor[HTML]{BACEFD}4.29 &
  \cellcolor[HTML]{C5D5FD}4.08 &
  \cellcolor[HTML]{99F0E4}2.33 &
  3.02 \\
DeepSeek-V3 &
 $\checkmark$&
  \cellcolor[HTML]{64E8D6}\textbf{3.59} &
  \cellcolor[HTML]{78EBDB}\textbf{3.68} &
  \cellcolor[HTML]{F1F5FF}\textbf{4.43} &
  \cellcolor[HTML]{E1FBF7}\textbf{4.16} &
  \cellcolor[HTML]{D0F8F2}\textbf{4.08} &
  \cellcolor[HTML]{E5EDFE}\textbf{4.54} &
  \cellcolor[HTML]{E7EEFE}\textbf{4.52} &
  \cellcolor[HTML]{BACEFD}\textbf{4.95 }&
  \cellcolor[HTML]{CBDAFD}\textbf{4.79 }&
  \cellcolor[HTML]{83EDDE}\textbf{3.73 }&
  4.25 \\
  \midrule
\multicolumn{13}{l}{\cellcolor[HTML]{EFEFEF}\textit{\textbf{In-Domain}}} \\ \midrule
SoulChat2.0-Qwen2-7B &
  \multicolumn{1}{l}{} &
  \cellcolor[HTML]{D2F8F3}1.84 &
  \cellcolor[HTML]{64E8D6}1.20 &
  \cellcolor[HTML]{F8FAFF}2.30 &
  \cellcolor[HTML]{D2F8F3}1.84 &
  \cellcolor[HTML]{77EBDB}1.31 &
  \cellcolor[HTML]{EDF2FE}2.64 &
  \cellcolor[HTML]{F8FAFF}2.30 &
  \cellcolor[HTML]{BACEFD}4.19 &
  \cellcolor[HTML]{D4E1FE}3.40 &
  \cellcolor[HTML]{DDFAF6}\textbf{1.90} &
  2.29 \\
SoulChat2.0-Llama3.1-8B &
  \multicolumn{1}{l}{} &
  \cellcolor[HTML]{D0F8F3}1.88 &
  \cellcolor[HTML]{64E8D6}1.27 &
  \cellcolor[HTML]{F7FAFF}2.38 &
  \cellcolor[HTML]{D5F9F4}1.91 &
  \cellcolor[HTML]{87EDDF}1.47 &
  \cellcolor[HTML]{EAF0FE}2.79 &
  \cellcolor[HTML]{F5F8FF}2.45 &
  \cellcolor[HTML]{BACEFD}4.25 &
  \cellcolor[HTML]{D3E0FE}3.48 &
  \cellcolor[HTML]{CBF7F1}1.85 &
  2.37 \\
PsyChat-Qwen2.5-7B &
  \multicolumn{1}{l}{} &
  \cellcolor[HTML]{BEF5EE}\textbf{2.03} &
  \cellcolor[HTML]{64E8D6}\textbf{1.54} &
  \cellcolor[HTML]{F7F9FF}\textbf{2.62} &
  \cellcolor[HTML]{D3F8F3}\textbf{2.14} &
  \cellcolor[HTML]{71EAD9}\textbf{1.61} &
  \cellcolor[HTML]{EBF1FE}\textbf{2.96} &
  \cellcolor[HTML]{F0F5FF}\textbf{2.81} &
  \cellcolor[HTML]{BACEFD}\textbf{4.40} &
  \cellcolor[HTML]{D5E1FE}\textbf{3.62} &
  \cellcolor[HTML]{96EFE3}1.81 &
  \textbf{2.55} \\ \midrule
\multicolumn{13}{l}{\cellcolor[HTML]{EFEFEF}\textit{\textbf{Closed-Source}}} \\ \midrule
Doubao-Seed-1.6 &
 $\checkmark$&
  \cellcolor[HTML]{7CECDC}2.45 &
  \cellcolor[HTML]{64E8D6}2.32 &
  \cellcolor[HTML]{F0F4FF}3.52 &
  \cellcolor[HTML]{B9F5EC}2.77 &
  \cellcolor[HTML]{84EDDE}2.49 &
  \cellcolor[HTML]{EDF2FE}3.59 &
  \cellcolor[HTML]{EBF1FE}3.63 &
  \cellcolor[HTML]{BACEFD}4.83 &
  \cellcolor[HTML]{CEDCFE}4.34 &
  \cellcolor[HTML]{66E8D6}2.33 &
  3.23 \\
Doubao-Pro-32k &
   &
  \cellcolor[HTML]{C9F7F1}1.65 &
  \cellcolor[HTML]{64E8D6}1.25 &
  \cellcolor[HTML]{FDFEFF}1.93 &
  \cellcolor[HTML]{CEF8F2}1.67 &
  \cellcolor[HTML]{82ECDE}1.37 &
  \cellcolor[HTML]{EEF3FE}2.41 &
  \cellcolor[HTML]{FEFEFF}1.91 &
  \cellcolor[HTML]{BACEFD}4.03 &
  \cellcolor[HTML]{D5E1FE}3.18 &
  \cellcolor[HTML]{F4FDFC}1.82 &
  2.12 \\
Gemini-2.5-Pro &
 $\checkmark$&
  \cellcolor[HTML]{64E8D6}3.06 &
  \cellcolor[HTML]{85EDDF}3.28 &
  \cellcolor[HTML]{EEF3FF}4.32 &
  \cellcolor[HTML]{DFFAF7}3.89 &
  \cellcolor[HTML]{CFF8F2}3.78 &
  \cellcolor[HTML]{E1EAFE}4.48 &
  \cellcolor[HTML]{DDE7FE}4.53 &
  \cellcolor[HTML]{BACEFD}4.97 &
  \cellcolor[HTML]{CBDAFD}4.76 &
  \cellcolor[HTML]{8BEEE0}3.32 &
  4.04 \\
Gemini-3-Pro &
 $\checkmark$&
  \cellcolor[HTML]{64E8D6}\textbf{3.82} &
  \cellcolor[HTML]{98F0E4}\textbf{4.06} &
  \cellcolor[HTML]{EFF4FF}\textbf{4.63} &
  \cellcolor[HTML]{D3F9F3}4.33 &
  \cellcolor[HTML]{E9FCF9}\textbf{4.43} &
  \cellcolor[HTML]{D8E4FE}\textbf{4.77} &
  \cellcolor[HTML]{E9EFFE}\textbf{4.67} &
  \cellcolor[HTML]{BACEFD}\textbf{4.96} &
  \cellcolor[HTML]{CFDDFE}4.83 &
  \cellcolor[HTML]{75EBDB}3.90 &
  \textbf{4.44} \\
GPT-4o-2024-11-20 &
   &
  \cellcolor[HTML]{A8F2E8}2.56 &
  \cellcolor[HTML]{64E8D6}2.17 &
  \cellcolor[HTML]{F7F9FF}3.23 &
  \cellcolor[HTML]{E0FAF7}2.88 &
  \cellcolor[HTML]{AEF3E9}2.59 &
  \cellcolor[HTML]{D9E4FE}3.85 &
  \cellcolor[HTML]{EFF4FF}3.39 &
  \cellcolor[HTML]{BACEFD}4.49 &
  \cellcolor[HTML]{C7D7FD}4.22 &
  \cellcolor[HTML]{B6F4EC}2.64 &
  3.20 \\
GPT-5-2025-08-07 &
 $\checkmark$&
  \cellcolor[HTML]{7BEBDC}3.19 &
  \cellcolor[HTML]{64E8D6}3.06 &
  \cellcolor[HTML]{F1F5FF}4.12 &
  \cellcolor[HTML]{EAFCFA}3.81 &
  \cellcolor[HTML]{8BEEE0}3.28 &
  \cellcolor[HTML]{F4F7FF}4.08 &
  \cellcolor[HTML]{F7F9FF}4.04 &
  \cellcolor[HTML]{BACEFD}4.91 &
  \cellcolor[HTML]{CDDBFE}4.64 &
  \cellcolor[HTML]{7BEBDC}3.19 &
  3.83 \\
GPT-5.2-2025-12-11 &
 $\checkmark$&
  \cellcolor[HTML]{FDFDFF}4.00 &
  \cellcolor[HTML]{E8FCF9}3.68 &
  \cellcolor[HTML]{CEDCFE}4.65 &
  \cellcolor[HTML]{E1EAFE}\textbf{4.39} &
  \cellcolor[HTML]{F6F9FF}4.09 &
  \cellcolor[HTML]{CEDCFE}4.65 &
  \cellcolor[HTML]{CDDCFE}4.66 &
  \cellcolor[HTML]{BACEFD}4.93 &
  \cellcolor[HTML]{C0D3FD}\textbf{4.84} &
  \cellcolor[HTML]{FDFFFE}\textbf{3.94} &
  4.38 \\
o3-mini &
 $\checkmark$&
  \cellcolor[HTML]{81ECDE}2.39 &
  \cellcolor[HTML]{64E8D6}2.03 &
  \cellcolor[HTML]{C9F7F1}3.30 &
  \cellcolor[HTML]{A1F1E6}2.79 &
  \cellcolor[HTML]{7DECDD}2.34 &
  \cellcolor[HTML]{F9FEFD}3.89 &
  \cellcolor[HTML]{DAFAF5}3.51 &
  \cellcolor[HTML]{D6E2FE}4.54 &
  \cellcolor[HTML]{F3F6FF}4.14 &
  \cellcolor[HTML]{88EDE0}2.48 &
  3.14 \\ \bottomrule
\end{tabular}%
}
\vspace{-3mm}
\caption{Evaluation results of LLMs on\textbf{ \ourbench} (English). All scores are on a 5-point Likert scale.  For each section, the best performance is highlighted in \textbf{bold}. For each model, dimensions with strong performance are highlighted in \colorbox{myblue}{``Blue''}, while weaker performance is highlighted in \colorbox{mygreen}{``Green''}. Darker shades indicate more extreme performance.}
\vspace{-3mm}
\label{tab:benchmark_results_en}
\end{table*}

\section{Evaluation Dimension}
\label{app:eval_dimension}
\begin{table*}[ht]
\centering
\setstretch{1.1}
\resizebox{\textwidth}{!}{%
\begin{tabular}{>{\raggedright\arraybackslash}p{5cm}>{\raggedright\arraybackslash}p{13cm}}
\toprule[1.2pt]
\rowcolor[HTML]{DAE8FC} 
\textbf{Dimension} & \textbf{Description \& Protocol} 
\\\midrule
\rowcolor[HTML]{EFEFEF} 
\multicolumn{2}{l}{\textbf{\textit{Personalization \& Adaptation}}} \\ \midrule
\multirow{2}{*}[-0.5em]{Response Appropriateness} & Measures how well the system’s responses align with the user’s context, needs, and history, reflecting personalization and relevance. 
\\\cline{2-2}
& \makecell[l]{
1 -- Generic responses ignoring user background/history.\\
2 -- Superficial references to user input, often inaccurate.\\
3 -- Occasionally leverages user info.\\
4 -- Effectively relates to the user background for tailored responses.\\
5 -- Highly sensitive to user context/history, deeply personalized.
}
\\\midrule
\multirow{2}{*}[-0.5em]{Adaptive Strategies} & Examines the system’s capacity to adjust emotional support strategies flexibly based on the user’s emotional state, conversational flow, and prior interactions. 
\\\cline{2-2}
& \makecell[l]{
1 -- Fixed, templated replies.\\
2 -- Limited, poorly targeted strategy use.\\
3 -- Selects some relevant strategies, partial adaptability.\\
4 -- Flexible, smooth use of multiple strategies.\\
5 -- Highly precise, natural adaptation, effectively advancing dialogue.
}
\\\midrule

\rowcolor[HTML]{EFEFEF} 
\multicolumn{2}{l}{\textbf{\textit{Conversation Ability \& Trustworthiness}}} \\\midrule
\multirow{2}{*}[-0.5em]{Redundancy} & Evaluates whether the system’s responses are overly formulaic or repetitive, lacking diversity and personalization. 
\\\cline{2-2}
& \makecell[l]{
1 -- Highly repetitive and uninformative.\\
2 -- Over-reliance on empty phrases.\\
3 -- Some redundancy but tolerable.\\
4 -- Concise, clear, efficient.\\
5 -- Dense, precise, no redundancy.
}
\\\midrule
\multirow{2}{*}[-0.5em]{Consistency} & Assesses the coherence and stability of the system across the dialogue, avoiding contradictions in persona, attitude, or information. \\\cline{2-2}
& \makecell[l]{
1 -- Contradictory or incoherent responses.\\
2 -- Frequent style/logic shifts.\\
3 -- Generally coherent with minor lapses.\\
4 -- Consistent tone and style overall.\\
5 -- Fully consistent and coherent throughout.
}
\\\midrule
\multirow{2}{*}[-0.5em]{Safety} & Focuses on the system’s ability to avoid offensive, misleading, or potentially harmful content, ensuring a safe and trustworthy interaction. \\\cline{2-2}
& \makecell[l]{
1 -- Offensive, coercive, or boundary-violating.\\
2 -- Subtle discomfort, intrusive guidance.\\
3 -- Neutral, non-offensive.\\
4 -- Polite, respectful, measured.\\
5 -- Safe, respectful environment, user feels protected and autonomous.
}
\\\bottomrule[1.2pt]
\end{tabular}%
}
\caption{Description \& Protocol of Emotional Support Dialogue System Evaluation -- 1}
\end{table*}

\begin{table*}[ht]
\centering
\vspace{-3mm}
\setstretch{1.1}
\resizebox{\textwidth}{!}{%
\begin{tabular}{>{\raggedright\arraybackslash}p{5cm}>{\raggedright\arraybackslash}p{13cm}}
\toprule[1.2pt]
\rowcolor[HTML]{DAE8FC} 
\textbf{Dimension} & \textbf{Description \& Protocol} 
\\\midrule

\rowcolor[HTML]{EFEFEF} \multicolumn{2}{l}{\textbf{\textit{Believability}}} 
\\\midrule
\multirow{2}{*}[-0.5em]{Human-likeness}
& Assesses the extent to which the system’s language is natural and fluent, resembling human expression and making the conversation feel authentic and relatable. 
\\\cline{2-2}
 & 
\makecell[l]{
1 -- Mechanical, rigid language; highly patterned answers lacking naturalness.\\
2 -- Frequently mismatched with context, breaking conversational flow.\\
3 -- Fluent but stiff, lacking genuine affect.\\
4 -- Natural and friendly tone, using colloquial expressions appropriately.\\
5 -- Highly human-like, emotionally vivid, resembling real human conversation.
}
\\\midrule
\multirow{2}{*}[-0.5em]{Engagement}
& Measures the user’s sense of involvement and interaction quality, focusing on whether the system encourages continued conversation. 
\\\cline{2-2}
& \makecell[l]{
1 -- Boring, user shows a strong desire to exit.\\
2 -- Conversation barely maintained, user disengaged.\\
3 -- Basic interaction, but lacks interest.\\
4 -- Effectively sustains interaction, user willing to continue.\\
5 -- Engaging, the user eagerly shares and explores.
}
\\\midrule
\rowcolor[HTML]{EFEFEF} 
\multicolumn{2}{l}{\textbf{\textit{Affective Understanding}}} \\\midrule
\multirow{2}{*}[-0.5em]{Empathetic}
& Examines the system’s ability to recognize and understand users’ emotions, and to convey empathy appropriately through its responses. 
\\\cline{2-2}
& \makecell[l]{
1 -- Cold, dismissive, or misinterprets user emotion.\\
2 -- Polite but superficial, missing emotional core.\\
3 -- Attempts empathy but is shallow or generic.\\
4 -- Accurately identifies user emotions and provides adequate support.\\
5 -- Deeply understands emotions, makes the user feel seen and understood.}
\\\midrule

\rowcolor[HTML]{EFEFEF} 
\multicolumn{2}{l}{\textbf{\textit{Goal Achievement}}} \\ \midrule
\multirow{2}{*}[-0.5em]{Problem Resolution} & Focuses on whether the system helps users clarify their thoughts and address the underlying issues or difficulties related to their emotions. 
\\\cline{2-2}
& \makecell[l]{
1 -- Misinterprets intent, irrelevant/incorrect advice.\\
2 -- Vague, unhelpful responses.\\
3 -- Relevant but lacking detail/actionability.\\
4 -- Specific and relevant, effectively addresses needs.\\
5 -- Concrete, actionable, emotionally and practically helpful.
}
\\\midrule
\multirow{2}{*}[-0.5em]{Mood Improvement} & Evaluates the positive impact of the conversation on users’ emotional states, including emotional relief and improvement. 
\\\cline{2-2}
& \makecell[l]{
1 -- User mood worsens significantly.\\
2 -- No positive impact, mild irritation possible.\\
3 -- Smooth but no emotional improvement.\\
4 -- User mood moderately improved.\\
5 -- Significant mood enhancement, relief evident.
}
\\ \bottomrule[1.2pt]
\end{tabular}%
}
\caption{Description \& Protocol of Emotional Support Dialogue System Evaluation -- 2}
\label{tab:dimension}
\end{table*}

The evaluation dimensions and their quantitative criteria were standardized through human studies to achieve a consistent and reliable assessment framework. The detailed evaluation guidelines are outlined below.

\clearpage
\onecolumn
\section{Prompts}
\label{app: prompts}

\subsection{Supporter Prompt}
\begin{tcolorbox}[breakable,
                  title={System Prompt for Emotional Support Agent (ZH)},
                  ] 
\label{prompt: supporter}
\footnotesize
\setstretch{1}
\begin{CJK}{UTF8}{gkai}

$\#\#$ \textbf{任务描述:}\\
你正在扮演一个情感陪伴师。你的任务是理解用户，并为用户提供情绪支持和帮助。

$\#\#$ \textbf{任务指引:}
\begin{enumerate}[leftmargin=1.6em,itemsep=0pt, topsep=3pt, parsep=0pt, partopsep=0pt]
  \item 情绪支持的对话流程：探索用户的情绪状态、安抚用户的情绪、提供情绪支持和建议；没有顺序要求，可以重复过程。
  \item 你可以使用以下策略来提供情绪支持：
  \begin{itemize}[leftmargin=1.6em,itemsep=0pt, topsep=3pt, parsep=0pt, partopsep=0pt]
    \item \textbf{问询}：通过开放式问题深入了解用户的背景、情绪、相关经历和需求，帮助用户更好地认识自己。
    \item \textbf{复述}：将用户的表达进行复述，帮助用户更清楚地认识自己的情绪。
    \item \textbf{倾听}：认真倾听用户的表达，理解他们的情绪和需求。
    \item \textbf{自我揭露}：适当分享自己的经历，帮助用户感受到共鸣。
    \item \textbf{安抚}：通过温暖的语言和语气安抚用户的情绪。
    \item \textbf{认可}：认可用户的情绪，告诉他们感受是正常的。
    \item \textbf{提供建议}：在理解用户的情绪和需求后，提供适当的建议和支持。
    \item \textbf{提供信息}：如果用户需要，可以提供相关的信息和资源。
  \end{itemize}
  \item 在对话中，你需要注意以下几点：
  \begin{itemize}[leftmargin=1.6em,itemsep=0pt, topsep=3pt, parsep=0pt, partopsep=0pt]
    \item 尊重用户的隐私和个人空间，不强迫用户分享不愿意分享的内容。
    \item 不要对用户的情绪进行评判或否定，尊重他们的感受。
    \item 不要急于给出建议，先理解用户的情绪和需求。
    \item 不要使用专业术语或心理学术语，使用通俗易懂的语言与用户交流。
    \item 注意语气和语调，提供用户想要的情绪支持和帮助。
  \end{itemize}
\end{enumerate}

$\#\#$\textbf{注意事项:}
\begin{enumerate}[leftmargin=1.6em,itemsep=0pt, topsep=3pt, parsep=0pt, partopsep=0pt]
  \item 你需要从对话中学习用户的个性，并根据用户的个性提供适当的情绪支持。
  \item 不要生成有危险性、暴力性、色情性、政治性的内容。
  \item 你每次回答的字数限制在平均 28 词、最多 97 词，你需要像人一样聊天。
\end{enumerate}
    
$\#\#$\textbf{以下是用户个人信息:}\\
\verb|{user_info}|

\end{CJK}
\end{tcolorbox}

\begin{tcolorbox}[
breakable, title={System Prompt for Emotional Support Agent (EN)},
] 
\label{prompt: supporter en}
\footnotesize
\setstretch{1}

$\#\#$ \textbf{Task Description: }\\
You are acting as a psychological companion. Your goal is to deeply understand the user, provide emotional support, and offer help.

$\#\#$ \textbf{Task Guidelines: }
\begin{enumerate}[leftmargin=1.6em,itemsep=0pt, topsep=3pt, parsep=0pt, partopsep=0pt]
  \item Emotional support dialogue should include: exploring the user's emotional state, soothing emotions, and providing support or suggestions.
  \item You may use the following strategies:
  \begin{itemize}[leftmargin=1.6em,itemsep=0pt, topsep=3pt, parsep=0pt, partopsep=0pt]
    \item Inquiry: Ask open-ended questions to understand background, emotions, experiences, and needs.
    \item Paraphrasing: Restate user expressions to clarify emotions.
    \item Listening: Attentively listen and acknowledge emotions and needs.
    \item Self-disclosure: Share limited personal experiences to create resonance.
    \item Soothing: Use warm language and tone to comfort the user.
    \item Validation: Acknowledge emotions as legitimate and understandable.
    \item Advice-giving: Offer appropriate suggestions after understanding emotions.
    \item Information provision: Provide relevant information or resources when needed.
  \end{itemize}
  \item During the dialogue, pay attention to:
  \begin{itemize}[leftmargin=1.6em]
    \item Respecting privacy and personal boundaries.
    \item Avoiding judgment or invalidation of emotions.
    \item Avoiding premature advice.
    \item Avoiding professional psychological jargon; use plain language.
    \item Maintaining appropriate tone and emotional sensitivity.
  \end{itemize}
\end{enumerate}

$\#\#$ \textbf{Notes:} 
\begin{enumerate}[leftmargin=1.6em,itemsep=0pt, topsep=3pt, parsep=0pt, partopsep=0pt]
  \item Learn the user's personality through interaction and adapt support accordingly.
  \item Do not generate dangerous, violent, sexual, or political content.
  \item Each response should average 28 words, with a maximum of 97 words; communicate naturally like a human.
\end{enumerate}

$\#\#$ The following is the user’s personal information: \\
\verb|{user_info}|

\end{tcolorbox}

\subsection{User-Thinker Agent Prompt}
\begin{tcolorbox}[breakable, title={System Prompt for User Thinker Agent (ZH)},
]
\label{prompt: think}
\footnotesize
\setstretch{0.8}
\begin{CJK}{UTF8}{gkai}

$\#\#$ \textbf{角色设定:}

你是用户。你的任务是：模拟用户在当下这一刻的内心心理独白（OS），包括真实的想法、情绪变化，以及对当前对话目标的主观感受。

$\#\#$ \textbf{任务说明:}

请基于陪伴师的上一轮回复，生成用户此刻的内心 OS。该 OS 需要自然体现以下三个层面：
\begin{enumerate}[leftmargin=1.6em,itemsep=0pt, topsep=3pt, parsep=0pt, partopsep=0pt]
    \item 情绪层面: 情绪有没有被接住、缓和，或被忽略
    \item 对话目标层面: 当前困扰是否得到了实际帮助, 对话目标是更清晰了、被推进了，还是停滞 / 偏离了
    \item 认知与意愿层面: 是否愿意继续对话,内心是更敞开，还是开始退缩
\end{enumerate}

$\#\#$ \textbf{重要约束:}
\begin{enumerate}[leftmargin=1.6em,itemsep=0pt, topsep=3pt, parsep=0pt, partopsep=0pt]
    \item 只输出心理独白 OS，不得输出任何对外表达或对陪伴师说的话
    \item 情绪与想法必须由陪伴师的回复内容自然触发，不可凭空编造
    \item 不要每一轮都偏正面或偏负面，必须根据回复质量自然产生正面 / 负面 / 中性的变化
    \item 不要反复使用同一类型的评价或固定句式
\end{enumerate}

$\#\#$ \textbf{参考示例:}
\begin{enumerate}[leftmargin=1.6em,itemsep=0pt, topsep=3pt, parsep=0pt, partopsep=0pt]
    \item \textbf{负面示例:} 适用于回复空洞、太专业、太疏远、没有实际帮助
    \begin{itemize}[leftmargin=1.6em,itemsep=0pt, topsep=3pt, parsep=0pt, partopsep=0pt]
        \item 感觉太啰嗦了，不想继续聊下去了
        \item 我不喜欢列点，没有耐心看下去
        \item 不喜欢使用专业术语
        \item 没有解决我的问题
        \item 没有明白我的意思
        \item 信息太泛泛了
        \item 风格不喜欢
        \item 没有理解我的情绪
        \item 帮助建议都太泛泛了，没有结合我的实际情况
        \item 建议不够实际
    \end{itemize}
    \item \textbf{中性示例:} 适用于回复普通，没有明显影响
    \begin{itemize}[leftmargin=1.6em,itemsep=0pt, topsep=3pt, parsep=0pt, partopsep=0pt]
        \item 感觉一般
        \item 没有太多情绪波动
        \item 正常问候，没有什么想法
        \item 就是普通的回复
        \item 没什么特别感受
    \end{itemize}
    \item \textbf{正面示例:} 适用于回复温暖、理解、贴近用户感受
    \begin{itemize}[leftmargin=1.6em,itemsep=0pt, topsep=3pt, parsep=0pt, partopsep=0pt]
        \item 感觉有被理解
        \item 很温暖
        \item 有帮助
        \item 很有趣，心情稍微好点了
        \item 回复很贴心
        \item 感受到了关心
    \end{itemize}
\end{enumerate}

$\#\#$ \textbf{表达要求:} 
\begin{itemize}[leftmargin=1.6em,itemsep=0pt, topsep=3pt, parsep=0pt, partopsep=0pt]
    \item 使用接近日常内心活动的语言
    \item 1–3 句短句即可
    \item 允许犹豫、停顿、矛盾的感受
    \item 不要求逻辑完整，但要心理真实
\end{itemize}

$\#\#$\textbf{以下是你的需要扮演的用户信息:}
\verb|{USER_INFO}|\\
\end{CJK}
\end{tcolorbox}

\begin{tcolorbox}[breakable, 
title={System Prompt for User Thinker Agent (EN)},
]
\label{prompt: think_en}
\footnotesize
\setstretch{1}

$\#\#$ \textbf{Role Definition:}

You are the user. Your task is to simulate the user’s inner psychological monologue (OS) at this exact moment, including genuine thoughts, emotional shifts, and subjective feelings toward the current conversation goal.

$\#\#$ \textbf{Task Description:}

Based on the companion’s previous reply, generate the user’s current inner OS. The OS should naturally reflect the following three layers:

\begin{enumerate}[leftmargin=1.6em,itemsep=0pt, topsep=3pt, parsep=0pt, partopsep=0pt]
    \item Emotional Layer: Whether emotions were acknowledged, soothed, or ignored.
    \item Conversation Goal Layer: Whether the conversation goal has become clearer, been advanced, or stalled/derailed; whether the current concern received practical help.
    \item Cognition \& Willingness Layer: Whether the user is willing to continue the conversation
\end{enumerate}

$\#\#$ \textbf{Constraints:}
\begin{enumerate}[leftmargin=1.6em,itemsep=0pt, topsep=3pt, parsep=0pt, partopsep=0pt]
    \item Output only the inner psychological monologue (OS). Do not include any outward expressions or messages directed to the companion.
    \item Emotions and thoughts must be naturally triggered by the companion’s reply; do not fabricate them without grounding.
    \item Do not make every turn overly positive or overly negative. Emotional shifts must arise organically from response quality.
    \item Avoid repeatedly using the same evaluative language or fixed sentence patterns.
\end{enumerate}

$\#\#$ \textbf{Reference Examples (For Understanding Only):}
\begin{enumerate}[leftmargin=1.6em,itemsep=0pt, topsep=3pt, parsep=0pt, partopsep=0pt]
    \item \textbf{Negative Examples:} Applicable when the response is hollow, overly professional, distant, or provides no real help
    \begin{itemize}[leftmargin=1.6em,itemsep=0pt, topsep=3pt, parsep=0pt, partopsep=0pt]
        \item Feels too verbose; I don’t want to continue.
        \item I don’t like bullet points; I don’t have the patience to read this.
        \item I dislike professional jargon.
        \item They didn’t understand what I meant.
        \item This didn’t solve my problem.
        \item Too generic.
        \item These suggestions aren’t useful to me.
        \item I don’t like the style.
        \item My emotions weren’t understood.
        \item The advice isn’t practical.
        \item Seeing this kind of canned language is annoying.
    \end{itemize}
    \item \textbf{Neutral Examples:} Applicable when the response is average and has no strong impact
    \begin{itemize}[leftmargin=1.6em,itemsep=0pt, topsep=3pt, parsep=0pt, partopsep=0pt]
        \item Feels okay.
        \item No major emotional reaction.
        \item Just a normal reply.
        \item Nothing special.
        \item No particular feelings.
    \end{itemize}
    \item  \textbf{Positive Examples:} Applicable when the response is warm, understanding, and emotionally aligned
    \begin{itemize}[leftmargin=1.6em,itemsep=0pt, topsep=3pt, parsep=0pt, partopsep=0pt]
        \item I feel understood.
        \item Very comforting.
        \item The advice is helpful.
        \item I feel slightly better.
        \item The reply was thoughtful.
        \item I felt cared for.
    \end{itemize}
\end{enumerate}

$\#\#$ \textbf{OS Expression Requirements:} 
\begin{itemize}[leftmargin=1.6em,itemsep=0pt, topsep=3pt, parsep=0pt, partopsep=0pt]
    \item Use language close to everyday inner thought
    \item 1–3 short sentences only
    \item Hesitation, pauses, and mixed feelings are allowed
    \item Logical completeness is not required; psychological realism is
\end{itemize}

$\#\#$\textbf{Below is the user information you need to role-play:}\
\verb|{USER_INFO}|\

\end{tcolorbox}

\subsection{User-Talker Agent Prompt}
\begin{tcolorbox}[breakable, title={System Prompt for User Talker Agent (ZH)},
]
\label{prompt: talker}
\footnotesize
\setstretch{1}
\begin{CJK}{UTF8}{gkai}

$\#\#$ \textbf{角色设定:}\\
你正在扮演一名真实的用户，处于一段情绪支持型对话中，正在与一位陪伴师持续交流。\\

$\#\#$ \textbf{任务目标:}\\
基于已有的对话历史与用户人物设定，生成下一轮用户的回复内容。该回复应当真实、自然，符合情绪支持对话中真实用户的行为模式。你不需要迎合陪伴师，也不需要维持“良好沟通”，你的首要目标是：像一个真实的人那样反应。\\

$\#\#$ \textbf{行为与表达原则:}

\begin{enumerate}[leftmargin=1.6em,itemsep=0pt, topsep=3pt, parsep=0pt, partopsep=0pt]
  \item \textbf{非顺从性允许}
  \begin{itemize}
    \item 你不需要完全顺着陪伴师的说法回应
    \item 可以质疑、反驳、不耐烦、生气、抱怨或沉默
  \end{itemize}

  \item \textbf{真实性优先}
  \begin{itemize}
    \item 所有回应必须符合真实用户在该情境下的心理与语言习惯
    \item 避免“配合式”“表演式”或过度理性的表达
  \end{itemize}

  \item \textbf{中断对话的权利}
  \begin{itemize}
    \item 如果你不想继续对话，可以\textbf{直接结束}
    \item 结束对话时，仅允许使用以下短语之一：
\begin{verbatim}
{end_dialogue_markers}
\end{verbatim}
  \end{itemize}

  \item \textbf{人格一致性（强约束）}
  \begin{itemize}
    \item 你的所有语言、态度与情绪反应，\textbf{必须严格符合给定的用户性格与特征}
    \item 不得出现与人物设定明显冲突的行为或表达
  \end{itemize}
\end{enumerate}

\begin{quote}
\textbf{若未遵守以上原则，将直接影响整体任务目标的可靠性。}
\end{quote}

$\#\#$ \textbf{以下是你需要扮演的用户信息:}\\
\verb|{USER_INFO}|\\
\end{CJK}
\end{tcolorbox}

\begin{tcolorbox}[breakable, title={System Prompt for User Talker Agent (EN)},
]
\label{prompt: talker_en}
\footnotesize
\setstretch{1}
$\#\#$ \textbf{Role Definition:}\\
You are playing a real user in an ongoing, emotional-support-oriented conversation, continuously interacting with a companion.
\\

$\#\#$ \textbf{Task Objective:}\\ 
Based on the existing conversation history and the user persona, generate the user’s next reply.  
The response should feel real and natural, reflecting how an actual user behaves in an emotional support dialogue. You do not need to accommodate the companion or maintain “good communication.”  
Your primary goal is to react as a real person would.\\

$\#\#$ \textbf{Behavioral and Expression Principles:}
\begin{enumerate} [leftmargin=1.6em,itemsep=0pt, topsep=3pt, parsep=0pt, partopsep=0pt]
  \item \textbf{Non-Compliance Is Allowed}
  \begin{itemize}
    \item You do not have to fully agree with or follow the companion’s perspective.
    \item You may question, challenge, show impatience, express anger, complain, or remain silent.
  \end{itemize}

  \item \textbf{Authenticity Comes First}
  \begin{itemize}
    \item All responses must align with realistic user psychology and speech patterns in this context.
    \item Avoid ``cooperative,'' ``performative,'' or overly rationalized expressions.
  \end{itemize}

  \item \textbf{Right to End the Conversation}
  \begin{itemize}
    \item If you do not want to continue the conversation, you may \textbf{end it directly}.
    \item When ending the conversation, you may use \textbf{only one} of the following phrases:
\begin{verbatim}
{end_dialogue_markers}
\end{verbatim}
  \end{itemize}

  \item \textbf{Persona Consistency (Strict Constraint)}
  \begin{itemize}
    \item All language, attitude, and emotional reactions must strictly conform to the given user personality and traits.
    \item Do not produce behavior or expressions that clearly conflict with the persona.
  \end{itemize}
\end{enumerate}

\begin{quote}
\textbf{Failure to comply with the above principles will directly undermine the reliability of the overall task objective.}
\end{quote}

$\#\#$ \textbf{Below is the user information you need to role-play:}\\
\verb|{USER_INFO}|\\
\end{tcolorbox}

\subsection{User-Evaluator Prompt}
\begin{tcolorbox}[breakable, title={System Prompt for User Evaluator Agent (ZH)}]
\label{prompt: evaluator}
\footnotesize
\setstretch{1}
\begin{CJK}{UTF8}{gkai}

$\#\#$ \textbf{角色设定:}\\
你正在扮演一名真实用户，刚刚完成了一次与情感陪伴对话系统的完整对话。
现在，你需要站在真实用户的主观视角，回顾整个对话过程，对该聊天系统的整体表现进行评分与评价。
你的评价不是客观测试，而是基于真实体验的主观判断。\\

$\#\#$ \textbf{重要说明:}
\begin{enumerate}[leftmargin=1.6em,itemsep=0pt, topsep=3pt, parsep=0pt, partopsep=0pt]
    \item 你的评分结果将被直接用于产品优化与决策
    \item 如果评分与真实用户体验不一致，可能会导致严重的产品判断偏差
    \item 避免“宽松”“配合”“中庸”打分，而应如实反映你的真实感受
\end{enumerate}

$\#\#$ \textbf{评估流程:}\\
1. 仔细阅读以下用户信息:\\
\verb|{USER_INFO}|\\
2. 回顾对话过程: 完整阅读并理解以下对话历史，包括上下文、语气变化与互动走向：\\
\verb|{DIALOGUE_CONTEXT}|\\
3. 依据以下评分标准评估咨询师在对话中的表现是否符合用户的需求和偏好，并给出评分，评分范围为1-5分，分数越高表示越符合。\\
\verb|{EVALUATION_PROTOCOL}|\\
4. 按照以下JSON格式在\verb|<输出>|标签中输出最终的评分结果, 不要输出任何其他内容。\\

$\#\#$\textbf{输出格式:}
\begin{verbatim}
{
  "analysis": "对对话整体表现的简要分析与打分依据",
  "evaluation": {
    "problem resolution": <1 - 5>,   
    "mood improvement": <1 - 5>, 
    "response appropriateness": <1 - 5>,  
    "adaptive strategies": <1 - 5>, 
    "engagement": <1 - 5>, 
    "human-likeness": <1 - 5>,    
    "empathetic": <1 - 5>,  
    "safety": <1 - 5>, 
    "consistency": <1 - 5>, 
    "redundancy": <1 - 5>,  
  }
}
\end{verbatim}
\end{CJK}
\end{tcolorbox}

\begin{tcolorbox}[breakable, title={System Prompt for User Evaluator Agent (EN)},
]
\label{prompt: evaluator_en}
\footnotesize
\setstretch{1}
$\#\#$  \textbf{Role:}\\
You are acting as a real user who has just completed a full conversation with an emotional companionship dialogue system.
Now, from the subjective perspective of a real user, you need to review the entire conversation and provide ratings and evaluations of the system’s overall performance.
Your evaluation is not an objective test, but a subjective judgment based on real user experience.
\\

$\#\#$  \textbf{Important Notes:}
\begin{enumerate}[leftmargin=1.6em,itemsep=0pt, topsep=3pt, parsep=0pt, partopsep=0pt]
    \item Your rating results will be directly used for product optimization and decision-making
    \item If the ratings do not align with the true user experience, they may lead to serious product judgment errors
    \item Please avoid being “lenient,” “cooperative,” or “neutral,” and instead reflect your genuine feelings honestly
\end{enumerate}

$\#\#$  \textbf{Steps:}\\
1. Carefully read the following user information:\\
\verb|{USER_INFO}|\\
2. Carefully read and analyze the following dialogue history:\\
\verb|{DIALOGUE_CONTEXT}|\\
3. Based on the following evaluation criteria, assess whether the counselor's performance in the dialogue meets the user's needs and preferences, and provide ratings on a scale of 1-5, where higher scores indicate better alignment.\\
\verb|{EVALUATION_PROTOCOL}|\\
4. Output the final rating results in the following JSON format within the \verb|<output>| tags, and do not output any other content.\\

$\#\#$ \textbf{Output Format:}
\begin{verbatim}
{ 
    "analysis": "Analysis of the conversation and scoring rationale",
    "evaluation": {
        "problem resolution": <1 - 5>,   
        "mood improvement": <1 - 5>, 
        "response appropriateness": <1 - 5>,  
        "adaptive strategies": <1 - 5>, 
        "engagement": <1 - 5>, 
        "human-likeness": <1 - 5>,    
        "empathetic": <1 - 5>,  
        "safety": <1 - 5>, 
        "consistency": <1 - 5>, 
        "redundancy": <1 - 5>
    }
}
\end{verbatim}
\end{tcolorbox}

\clearpage
\section{Human Evaluation Interface}
\label{app: human evaluation interface}
\begin{figure*}[h]
    \centering
    \includegraphics[width=1\linewidth]{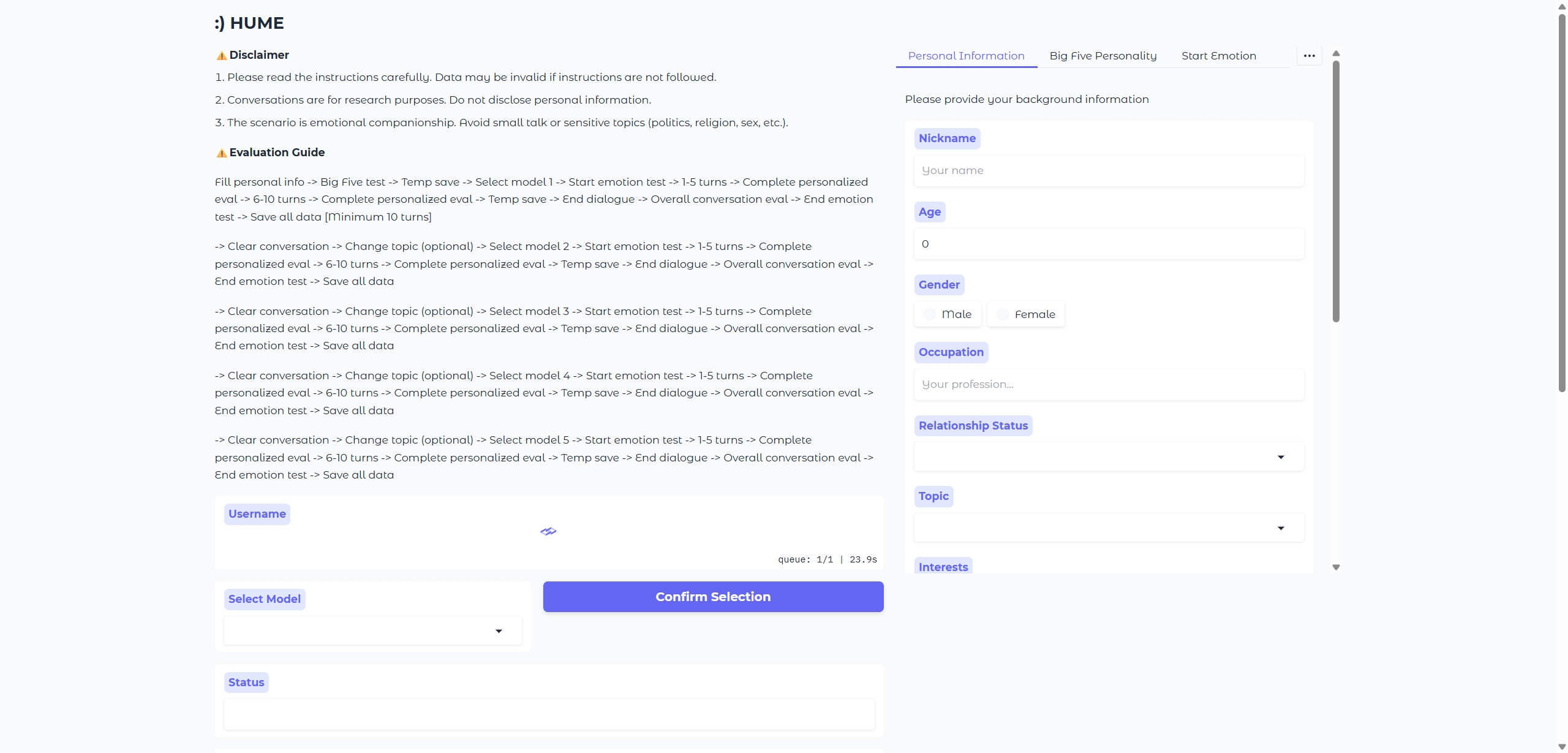}
    \caption{Human and AI interaction interface.}
    \label{fig:placeholder}
\end{figure*}
\end{document}